\documentclass[journal, twoside]{IEEEtran}

\usepackage[colorlinks=true,
    linkcolor=blue,   
    citecolor=blue,    
    urlcolor=black ]{hyperref}
\usepackage{url}
\usepackage{graphics} 
\usepackage{epsfig} 
\usepackage{mathptmx} 
\usepackage{times} 
\usepackage{amsmath} 
\usepackage{amssymb}  
\usepackage[noadjust]{cite}
\usepackage{multirow}
\usepackage{booktabs}
\usepackage{subfigure}
\usepackage{tabularx}
\usepackage{threeparttable}
\usepackage{indentfirst}
\usepackage{algpseudocode}
\usepackage{algorithmicx,algorithm}

\usepackage{makecell}
\usepackage{xcolor} 
\usepackage{soul}
\usepackage{comment}
\usepackage{enumitem}
\usepackage[normalem]{ulem}
\algdef{SE}[DOWHILE]{Do}{doWhile}{\algorithmicdo}[1]{\algorithmicwhile\ #1}

\begin{document}

\markboth{IEEE TRANSACTIONS ON ROBOTICS}{Liu \MakeLowercase{\textit{et al.}}: Automatic Tissue Traction Using Miniature Force-Sensing Forceps for Minimally Invasive Surgery}
%
\title{Automatic Tissue Traction Using Miniature Force-Sensing Forceps for Minimally Invasive Surgery}
\author{Tangyou~Liu, \textit{Graduate Student Member, IEEE}, Xiaoyi~Wang, Jay~Katupitiya, \\Jiaole~Wang$^{\ast}$, \textit{Member, IEEE}, and~Liao~Wu$^{\ast}$, \textit{Member, IEEE}
	\thanks{Manuscript received 23 June 2024; revised 1 September 2024; accepted 3 October 2024. This article was recommended for publication by Associate Editor E. De Momi and Editor A. Menciassi upon evaluation of the reviewers’ comments.
 This work was partly supported by Australian Research Council under Grant DP210100879, Heart Foundation under Vanguard Grant 106988, and UNSW Engineering under GROW Grant PS69063 awarded to Liao~Wu, partly by the Talent Recruitment Project of Guangdong under Grant 2021QN02Y839 and the Science and Technology Innovation Committee of Shenzhen under Grant JCYJ20220818102408018 \& GXWD20231129103418001  awarded to Jiaole Wang, and partly by the Tyree IHealthE PhD Top-Up Scholarship and China Scholarship Council PhD Scholarship awarded to Tangyou~Liu. $^{\ast}$Corresponding authors: Jiaole Wang \textit{(wangjiaole@hit.edu.cn)} and Liao Wu (\textit{liao.wu@unsw.edu.au}).}
	\thanks{Tangyou~Liu, Jay Katupitiya, and Liao~Wu are with the School of Mechanical \& Manufacturing Engineering, University of New South Wales, Sydney, NSW 2052, Australia.}
	\thanks{Xiaoyi~Wang is with the School of Electrical Engineering and Telecommunications, University of New South Wales, Sydney, NSW 2052, Australia.}
 \thanks{Jiaole Wang is with the School of Biomedical Engineering and Digital Health, Harbin Institute of Technology, Shenzhen, 518055, China.}
 \thanks{This article has supplementary material provided by the authors and color versions of one or more figures available at \href{https://doi.org/10.1109/TRO.2024.3486177}{https://doi.org/10.1109/TRO.2024.3486177}.}
 \thanks{Digital Object Identifier 10.1109/TRO.2024.3486177.}
}

\definecolor{d_c}{RGB}{0,0,250}

\maketitle

\begin{abstract}
A common limitation of autonomous tissue manipulation in robotic minimally invasive surgery (MIS) is the absence of force sensing and control at the tool level.
Recently, our team has developed miniature force-sensing forceps that can simultaneously measure the grasping and pulling forces during tissue manipulation.
Based on this design, here we further present a method to automate tissue traction that comprises grasping and pulling stages. 
During this process, the grasping and pulling forces can be controlled either separately or simultaneously through force decoupling.
The force controller is built upon a static model of tissue manipulation, considering the interaction between the force-sensing forceps and soft tissue.
The efficacy of this force control approach is validated through a series of experiments comparing targeted, estimated, and actual reference forces. 
To verify the feasibility of the proposed method in surgical applications, various tissue resections are conducted on \textit{ex vivo} tissues employing a dual-arm robotic setup. 
Finally, we discuss the benefits of multi-force control in tissue traction, evidenced through comparative analyses of various \textit{ex vivo} tissue resections with and without the proposed method, and the potential generalization with traction on different tissues.
The results affirm the feasibility of implementing automatic tissue traction using miniature forceps with multi-force control, suggesting its potential to promote autonomous MIS.
A video demonstrating the experiments can be found at \href{https://youtu.be/f5gXuXe67Ak}{https://youtu.be/f5gXuXe67Ak}.  
\end{abstract}

\section{Introduction}
\label{Section:introduction}

\IEEEPARstart{S}{urgical} robots equipped with miniature instruments have been extensively studied in recent years, demonstrating significant potential for minimally invasive surgery (MIS) \cite{yang2018grand,dupont2021decade,dupont2022continuum,razjigaev2022end,nwafor2023design,wu2017development}.
Within this field, miniature forceps are increasingly employed for delicate tissue manipulation \cite{wang2023novel, hu2023design, price2023using, gao2023transendoscopic, wu2022camera, kong2022design, yu2016development}.
In addition, as surgical robots advance towards enhanced autonomy, the importance of precise robotic tissue manipulation is increasingly recognized \cite{yang2018grand,li2023design}.
To facilitate efficient and safe tissue manipulation in robotic MIS, accurate control over the grasping and pulling forces exerted by miniature forceps is vital, which is amplified by the risk of severe complications associated with improper manipulation forces \cite{van2009value,khadem2016modular,abiri2019multi,skandalakis2021surgical,liu2022recent}.
However, this critical functionality remains absent in most systems employing miniature instruments.

\begin{figure*}[t!] 	
	\centering 
	\includegraphics[width=1.0\textwidth]{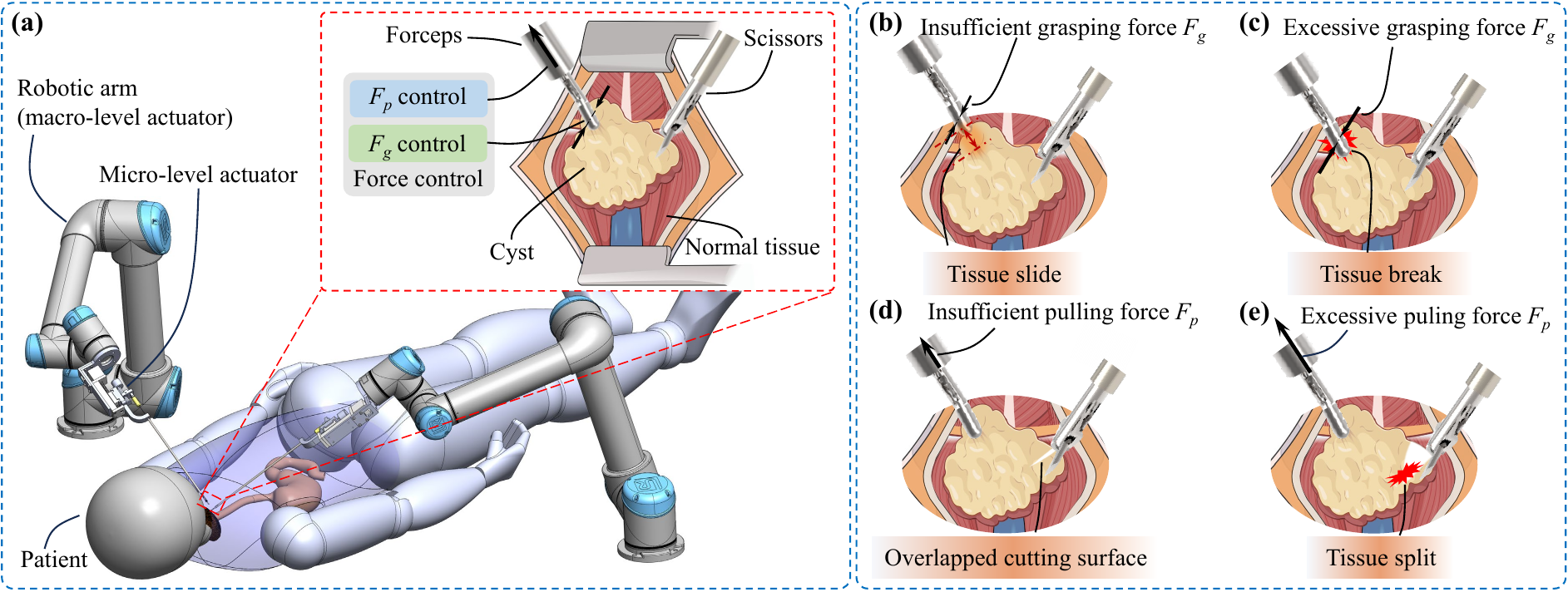} 	
	\caption{Typical tissue resection with forceps and scissors.
	(a) Dual-arm robotic system for lesion resection, where the forceps are responsible for tissue traction, while the scissors are responsible for tissue cutting (inset adapted from \cite{skandalakis2021surgical}).
    The grasping force $F_g$ control and pulling force $F_p$ control are crucial for efficient and safe tissue traction.
	(b) and (c) show tissue slide and break results from insufficient and excessive grasping forces $F_g$, respectively. 
	(d) insufficient tissue pulling force $F_p$ results in cutting surface overlapping and prevents the scissors from performing continuous cuts.
	(e) unexpected tissue split results from excessive pulling force $F_p$. 
 } 
	\label{Figure:Introduction} 
\end{figure*}

Tissue traction, comprising grasping and pulling, is a frequently performed surgical task \cite{skandalakis2021surgical}.
For example, Fig.~\ref{Figure:Introduction}(a) illustrates a robotic thyroidectomy scenario where forceps and scissors collaboratively undertake cyst dissection. During such procedures, the traction executed by the forceps necessitates both separate and simultaneous control of grasping and pulling forces. 
This control is crucial to prevent tissue damage and ensure optimal surgical outcomes, as shown in Fig.~\ref{Figure:Introduction}(b)-(e).
In particular, grasping and pulling forces are highly coupled during the tissue pulling process, and they are significantly influenced by the interaction between instrument and tissue \cite{kim2018sensorized, seok2019compensation, liu2023hapticsenabled}.
In response to the need for enhanced precision in robotic surgery, this paper introduces a method for automatic tissue traction using miniature forceps with controlled forces.

\subsection{Background}
To effectively control the forces exerted by forceps in surgery, accurate sensing of interaction forces is essential. 
Therefore, a large body of work has focused on techniques to enable force sensing in tissue manipulation, ranging from observing tissue deformation to developing tool-level force sensors.

Tissue deformation observation often relies on imaging techniques \cite{nazari2021image}. 
For instance, Haouchine \textit{et al.} employed endoscopic images to develop an underlying biomechanical model for estimating contact forces \cite{haouchine2018vision}. 
Beyond traditional tissue modeling, research has explored using learning methods to estimate interaction forces based on visual information. 
Lee \textit{et al.} achieved contact force estimation using a neural network-based method that integrates actuators’ electrical currents with camera images to enhance accuracy \cite{lee2018interaction}. 
However, these studies primarily concentrate on contact force estimation, often overlooking grasping forces, which are equally crucial in tissue manipulation.
Estimating grasping forces poses additional challenges, particularly due to the difficulty in visually monitoring and measuring tissue deformation between the forceps' jaws \cite{li2020super}.
Furthermore, these learning-based methods generally necessitate complex training processes and are difficult to prepare in practical surgery \cite{nazari2021image,nwoye2022artificial}.
The reliability of these methods also depends on precise, real-time modeling of tissue biomechanical properties, posing significant challenges in robustness and applicability across diverse surgical scenarios \cite{lanir2017multi}.

Another approach for force sensing in tissue manipulation is the integration of sensors directly onto surgical forceps \cite{bandari2019tactile}. 
For example, King \textit{et al.} \cite{king2009tactile} mounted two FlexiForce sensors on the forceps' jaws to enable tactile feedback and demonstrated its effectiveness in reducing grasping forces during operations.
Beyond directly utilizing commercial sensors, investigations were conducted on developing sensing modules tailored to customized forceps.
For instance, Lai \textit{et al.} \cite{lai2021three} integrated Fiber Bragg grating (FBG) sensors into forceps to enable pulling force sensing.
Using the same principle, Zarrin \textit{et al.} \cite{zarrin2018development} created a gripper capable of measuring grasping and axial forces with FBGs embedded into the gripper's jaws.
Capacitance-based sensors \cite{kim2015force,seok2019compensation} have also been installed in the forceps' jaws to achieve similar outcomes.
However, the significant size of these sensing modules and the extensive modifications required on the jaws hinder their integration into miniature forceps, such as those used in \cite{wu2022camera, kong2022design, wang2023novel, hu2023design, price2023using, gao2023transendoscopic}.
To overcome this challenge, our previous work \cite{liu2023hapticsenabled} introduced a novel vision-based module with a diameter of 4 mm that can be seamlessly embedded in miniature forceps, successfully achieving multi-modal force sensing (both grasping and pulling forces) in delicate tissue manipulation.

Despite these advances in force sensing techniques, how to use these capabilities for precise force control with miniature forceps in tissue manipulation has not received sufficient investigation and remains an open question.
Xiang~\textit{et al.}~\cite{xiang2023learning} adopted a learning-based approach to facilitate contact force estimation and control at the tip of a continuum robot.
Although the training of the neural networks involved only the actuator’s state and a reference contact force, the implementation phase necessitated prior biomechanical information.
In parallel, Wijayarathne \textit{et al.} \cite{wijayarathne2023real} developed a model to detect contact in force-modulated manipulation, employing Young's modulus of both the tool-tip and the contact surface.
In a different approach, Sheng \textit{et al.} \cite{sheng2021hybrid} bypassed the intricate modeling by directly affixing a force sensor at the instrument’s base, thus enabling direct control over the instrument-tissue contact force. 
However, it is important to note that the methodologies mentioned above were all developed for probe-like instruments where only the contact forces are concerned.
None of them are directly applicable to surgical forceps which also involve grasping forces.
Khadem \textit{et al.} \cite{khadem2016modular} achieved control over the grasping forces of forceps on rubber by managing the driving force through a force sensor mounted at the proximal end of the tool. 
However, the study was limited to grasping forces only and neglected the control of pulling forces, which are also critical in delicate tissue manipulation.

In summary, despite the achievements mentioned above, effective control of the miniature forceps’ grasping and pulling forces remains largely unexplored.
In particular, to the best of the authors' knowledge, no study has investigated the simultaneous control of grasping and pulling forces in tissue manipulation, which poses a grand challenge due to tight force coupling \cite{kim2018sensorized, seok2019compensation, liu2023hapticsenabled}.
It is imperative to address this challenge to equip the miniature forceps with an essential capability for automatic tissue manipulation in MIS \cite{dupont2021decade,li2023design}.

\begin{figure}[t] 	 	
    \centering  	
    \includegraphics[width=0.488\textwidth]{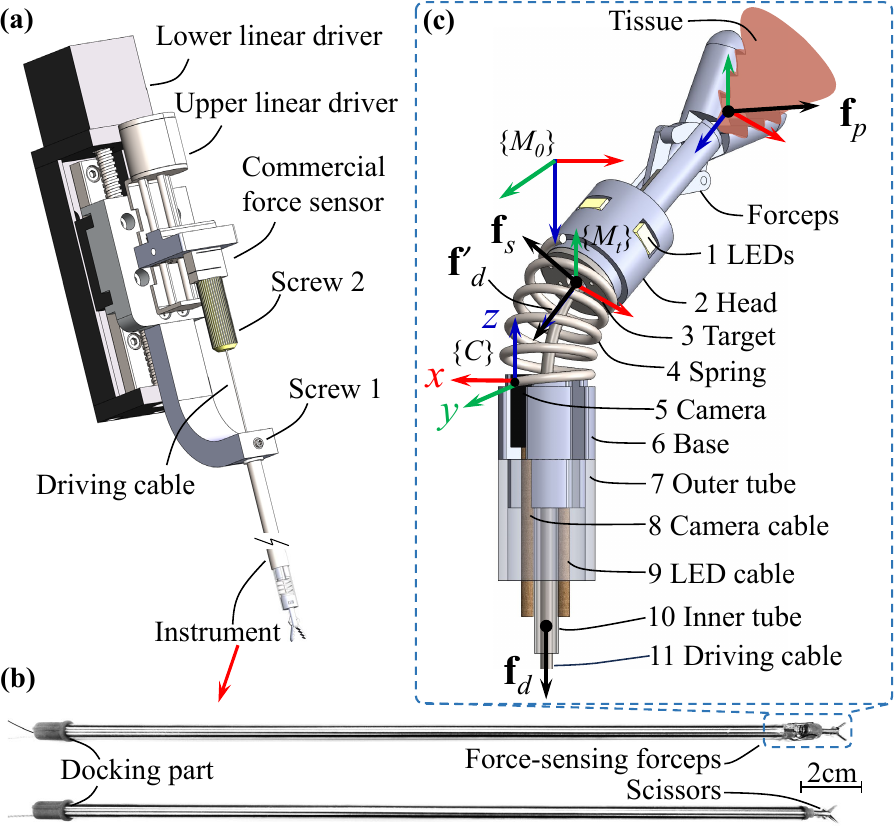}  	
    \caption{Portable instruments and micro-level actuator.
     (a) The micro-level actuator updated from \cite{liu2023hapticsenabled}, wherein the screw~1 is used to fix the instrument’s stainless tube, and the screw~2 is used to lock the forceps’ driving cable.
     (b) Instrument prototypes that integrate with the force-sensing forceps and scissors, wherein the docking part positions and orientates the instrument.
    (c) force-sensing forceps with a vision-based force sensing module presented in \cite{liu2023hapticsenabled}, where the camera (5) is used to measure the deformation of the spring (4) by tracking and estimating the pose of the target (3).
    } 	
    \label{Figure:Instrument} 
\end{figure} 

\subsection{Contribution}
This work is a substantial extension of our previous work \cite{liu2023hapticsenabled}, which presented a pair of miniature forceps capable of sensing grasping and pulling forces. 
Leveraging the sensorized forceps, in this article, we propose a novel method to enable the control of multiple forces and automatic tissue traction.
The main contributions of this article are as follows:
\begin{enumerate}
    \item A controller is proposed to manage and decouple the grasping and pulling forces of miniature forceps, enabling more precise tissue manipulation.
    \item Based on the force-controlled miniature forceps, automatic tissue traction is realized and implemented in \textit{ex vivo} tissue resections.
\end{enumerate}

We start by deriving a model of tissue manipulation that considers the interaction between forceps and soft tissues and analyzing its controllability and observability.
Then, we design a controller capable of regulating the grasping and pulling forces separately or simultaneously using the miniature forceps' multi-modal force sensing.
The efficacy of the force controller is validated through a series of experiments comparing the target, estimated, and reference results.

We further investigate the feasibility of automatic tissue traction based on the developed controller.
The automatic tissue traction includes two stages, grasping and pulling.
The grasping stage relies on a controlled grasping force, while the pulling stage is under the guidance of a controlled pulling force, as depicted in Fig.~\ref{Figure:Introduction}(a).
Integrating the force-sensing forceps in a dual-arm robotic system, we demonstrate the potential surgical application of automatic tissue traction by a series of resections on \textit{ex vivo} tissues.

Finally, we discuss the benefits of multi-force control in tissue traction, evidenced through comparative analyses of various \textit{ex vivo} resections with and without the proposed method, and the potential generalization with traction on different tissues.

\section{System Description and Modeling }
\label{system_modeling}

\begin{figure}[t] 	
	\centering 
    \includegraphics[width=0.488\textwidth]{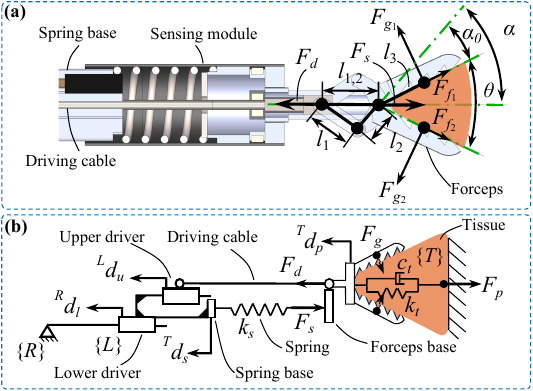} 
	\caption{ (a) Forces applied to the forceps when they grasp and pull a tissue include driving force $F_d$ from the cable, supporting force $F_s$ from the sensing module, contact force $F_g$ (also called grasping force), and friction force $F_f$ from the tissue. 
	(b) Model of tissue traction with force-sensing microforceps. ${}^{T}d_p$ and ${}^{T}d_s$ denote the displacement of the forceps’ jaws related to the tissue body \{$T$\} and that of the spring base related to \{$T$\}, respectively.
    ${}^{R}d_l$ is the lower driver displacement associated with its mounting points on the robotic arm \{$R$\}, while ${}^{L}d_u$ is the upper driver displacement related to its mounting points on the lower driver \{$L$\}.
  }
	\label{Figure:Modeling}
\end{figure}

\subsection{Force Sensing of Miniature Forceps}
The miniature forceps and their vision-based force estimation method have been presented in our previous work \cite{liu2023hapticsenabled}.
Integrated with a micro-level actuator, the efficacy of this new kind of miniature forceps has been validated for force estimation in robotic tissue manipulation tasks.
To reduce duplication without compromising comprehensibility, we briefly describe the prototype in this section. 
Subsequently, we clarify how the driving force and supporting force applied to the forceps are independently estimated, followed by the calculation of the pushing/pulling force and grasping force.

The micro-level actuator (Fig.~\ref{Figure:Instrument}(a)) is optimized by a portable design considering sterilization and delivery. 
The instrument integrated with the miniature force-sensing forceps (Fig.~\ref{Figure:Instrument}(b) shows the prototype) can achieve one-step mating with the micro-level actuator through a docking part and then be fixed by two screws.
To avoid interference from the camera and LED cables, an inner tube is concentrically installed along with the outer stainless tube to enclose the forceps’ driving cable only, as shown in Fig.~\ref{Figure:Instrument}(c).

The forces applied to the forceps when they push or pull a tissue are also shown in Fig.~\ref{Figure:Instrument}(c).
$\mathbf{f}_d$ is the driving force from the cable, $\mathbf{f}_p$ is the pushing/pulling force from the interactive tissue, and $\mathbf{f}_s$ is the supporting force applied to the forceps’ base from the installed sensing module.
The driving force $\mathbf{f}_d$ is directly measured by a commercial gauge force sensor installed at the proximal end of the instrument shown in Fig.~\ref{Figure:Instrument}(a).
The supporting force $\mathbf{f}_s$ is equal to the elastic force from the spring and can be formulated as
\begin{equation}
	\mathbf{f}_s = \mathbf{K}_s \,{} \mathbf{t}_s \, ,
 \label{equation:fs}
\end{equation}
where $\mathbf{K}_s$ is the spring’s stiffness matrix, $\mathbf{t}_s$ is the translation vector of ${}^{M_0}_{M_t}\mathbf{T}$, representing the spring’s deformation, and estimated by the sensing module’s camera by tracking the installed target.

Because the estimation of $\mathbf{f}_s$ solely depends on the spring’s deformation, it can be directly and independently estimated by the vision-based sensing module. 
Consequently, this allows for the driving force $\mathbf{f}_d$ and supporting force $\mathbf{f}_s$ applied to the forceps to be measured independently and concurrently.

According to the force equilibrium, the pulling force $\mathbf{f}_p$ can be obtained by
\begin{equation}
	\mathbf{f}_p = -{}^{M_0}_{M_t}\mathbf{T}\; \mathbf{f}_d \, - \mathbf{f}_s \,.
	\label{equation:fp} 
\end{equation}
We define the contact force between the tissue and forceps’ jaws as the grasping force $F_g$, as depicted in Fig.~\ref{Figure:Modeling}(a).
For convenience, we adopt $F_{\bullet}$ and $t_{\bullet}$ to denote the magnitude of force $\mathbf{f}_{\bullet}$ and displacement $\mathbf{t}_{\bullet}$ in the following sections.
Then, the grasping force $F_g$ can be formulated as
\begin{equation}
	F_g =  \frac{F_d \, l_2 \sin{\alpha}}{2l_3}\, ,
	\label{equation:grasping_force}	
\end{equation}
where 
\begin{equation} 		
    \alpha = \alpha_0 + \theta = \arccos \frac{{l^{2}_{1,2}} + l_2^2 - l_1^2}{2{l_1}\, l_{1,2}} \, .		
    \label{equation:alpha} 	
\end{equation}
The details of the grasping and pulling forces estimation can be found in \cite{liu2023hapticsenabled}.

\subsection{Model of Tissue Traction with Forceps}
\label{Section:Controller_software}
Force-sensing forceps pave the way for tissue manipulation, including controlling the grasping force $F_g$ and the pulling force $F_p$.
To facilitate efficient and safe robotic tissue traction, it is essential to control these multiple forces applied to the tissue.  

To establish the model of tissue traction using forceps, accurately estimating the interaction forces between the soft tissue and the forceps is essential.
Ideally, this force estimation can be achieved by constructing a tissue model, such as using the Kelvin–Voigt model \cite{stoll2006force, yu2007comparison}.
However, the diverse and complex characteristics of different tissues, coupled with their dynamic changes during the operation, pose significant challenges to precise modeling \cite{lanir2017multi}.
In the development of force-sensing forceps \cite{liu2023hapticsenabled}, we have enabled the real-time estimation of both grasping and pulling forces. 
Consequently, our system does not require a precise tissue model for force estimation.

In this study, to assess the observability and controllability of the tissue manipulation using miniature forceps, we simplify the grasped tissue by a Kelvin–Voigt model that consists of a viscous damper $c_t$ and elastic spring $k_t$ for the mathematical model derivation.
Assuming the forceps grasp the tissue stably without any slippage or tissue damage, we introduce a static model when the forceps grasp and pull the tissue, as depicted in Fig.~\ref{Figure:Modeling}(b).

The system has two inputs $\mathbf{u}= [u_1, \, u_2]^\top=[^{R}\dot d_l,\,  ^{L}\dot d_u ]^\top$, and two outputs $\mathbf{y} = [y_1, \, y_2]^\top = [F_g, \, F_p]^\top$.
According to Fig.~\ref{Figure:Modeling}(b), we have below two function groups:
\begin{align}
	\left\{
	\begin{array}{l}
        F_d = k_t \, {^{T}{d}_p} + c_t \, {^{T}\dot{d}_p} + k_s( {^{T}{d}_p} - {^{T}d_s}) \vspace{1ex}  \\ 
        F_p = k_t \, {^{T}{d}_p} + c_t \, {^{T}\dot{d}_p} 
	\end{array}\right. \, ,
	\label{Equation:output_1}
\end{align}
where ${^{T}\dot{d}_p} - {^{T}\dot{d}_s} = {^{T}\dot{t}_s}$, $^{T}d_p$ and $^{T}d_s$ denote the displacement of the forceps’ jaws related to the tissue body \{$T$\} and that of the spring base related to \{$T$\}.
\begin{align}
	\left\{
	\begin{array}{l}
        {^{T}d_p} = {^{R} d_l} + {^{L} d_u}  \\ 
        {^{T}d_s} = {^{R} d_l}
	\end{array}\right. \, ,
	\label{Equation:motor2tissue}
\end{align}
where $^{R}d_l$ is the lower driver displacement related to its mounting point on the robotic arm \{$R$\}, while $^{L}d_u$ is the upper driver displacement related to its mounting point on the lower driver \{$L$\}.

Then, defining the state variables as $\mathbf{x}=[x_1, \, x_2]^\top = [{}^{T}d_p,\,{}^{T}d_s]^\top$, we can formulate the state-space equations as
\begin{equation}
	\dot{\mathbf{x}}  = \mathbf{A} \mathbf{x} + \mathbf{B} \mathbf{u}\, ,
\end{equation}
with $\mathbf{A}=\mathbf{0}$ and 
$\mathbf{B}= 
\begin{bmatrix}
    {1}&{1}\\
    {1}&{0}
\end{bmatrix}$.

\noindent
Referring to (\ref{equation:grasping_force}), (\ref{Equation:output_1}), and (\ref{Equation:motor2tissue}), we can obtain the outputs as
\begin{equation}
    \mathbf{y} = \mathbf{C} \mathbf{x} + \mathbf{D} \mathbf{u}\, ,
\label{Equation:output}
\end{equation}
where
\[
{\mathbf{C} = 
\begin{bmatrix}
\frac{{l_2}\sin\alpha}{2{l_3}}(k_t + k_s) & - \frac{{l_2}\sin\alpha}{2{l_3}}k_s	 \\
    k_t & 0
\end{bmatrix}},
{\mathbf{D} = 
\begin{bmatrix}
	\frac{{l_2}\sin\alpha}{2{l_3}}c_t & \frac{{l_2}\sin\alpha}{2{l_3}}c_t\\
    c_t & c_t
\end{bmatrix}.}
\]
\noindent
Because $rank([\mathbf{B}, \,\mathbf{AB}]) = 2$ and $rank([\mathbf{C}^\top,\mathbf{A}^\top\mathbf{C}^\top]^\top) = 2$, the system is both controllable and observable.

\section{Methods for Automatic Tissue Traction}
To achieve precise and automatic tissue traction, the forceps must be capable of following predefined grasping and pulling force profiles. 
Consequently, it becomes essential to design controllers for grasping and pulling processes.
However, it is challenging to model tissue mechanical properties precisely in real time, and the controller needs to be adaptive to deal with various tissues.
Here, we adopt a standard and effective proportional–integral–derivative (PID) controller to enable the forceps to track these force profiles, 
\begin{equation}
	u(t) = {K_P}{\,}e(t) + {K_I}\int_0^t  {e(\tau){\,}d\tau}  + {K_D}{\,} \dot e(t) \, ,
 \label{Equation:PID_controller}
\end{equation}
where $e(t) = F^\ast_{\bullet}(t) - F^e_{\bullet}(t)$ is the error between the target force $F^\ast_{\bullet}(t)$ and the estimated value $F^e_{\bullet}(t)$, and $\bullet$ denotes subscript \{$g$, $p$\} for grasping and pulling forces, respectively.

\begin{figure}[t!] 	
	\centering 
	\includegraphics[width=0.488\textwidth]{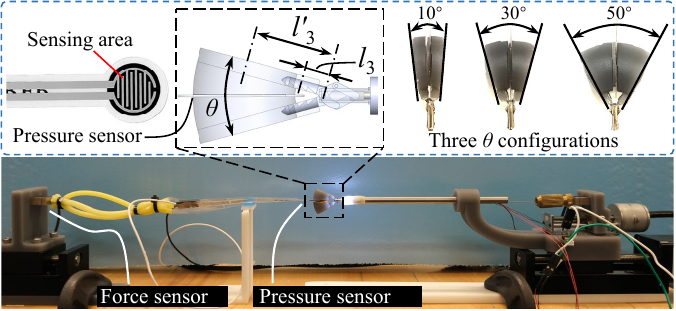} 
    	\caption{Experimental setup for force control evaluation, where a commercial gauge force sensor and pressure sensor were adopted to measure reference pulling and grasping forces, respectively.
	The inset shows three 3D-printed extension jaws configured with $\theta$ = $10^\circ$, $30^\circ$, and $50^\circ$.
 }
	\label{Figure:Force_setup}
\end{figure}

\subsection{Tissue Grasping with Controlled $F_g$}
\label{section:grasping_control}

\subsubsection{Controller Design}
In our forceps, the current $F^e_g$ is estimated and observed in each running cycle via (\ref{equation:grasping_force}).
Originally, we plan to apply a group of parameters $[K_{P,\: 1}, \: K_{I,\: 1}, \: K_{D,\: 1}]$ with controller (\ref{Equation:PID_controller}) to control the grasping force $F_g$.
However, from (\ref{equation:grasping_force}), we find that $F_g$ is positively related to $\theta$.
Therefore, we set the corresponding parameters $[K^{\theta}_{P,\: 1},\: K^{\theta}_{I,\: 1}, \: K^{\theta}_{D,\: 1}]$ of different $\theta$ configurations as below (\ref{equation:k_1_theta}) to minimize the difference in response.
\begin{equation}
    [K^{\theta}_{P,\: 1},\: K^{\theta}_{I,\: 1}, \: K^{\theta}_{D,\: 1}] = f(\theta) \, [K_{P,\: 1}, \: K_{I,\: 1}, \: K_{D,\: 1}] \, ,
    \label{equation:k_1_theta}
\end{equation}
where $f(\theta)=\frac{\sin{\alpha_0}}{\sin(\frac{\theta}{2}+\alpha_0)}$ is a function related to $\theta$.

In addition, considering the deformation of the spring and the press deformation of the tissue, we set the speed of the lower driver $^{R} \dot d_l$ as 
\begin{equation}
    ^{R} \dot d_l(t) = -\frac{^{T}{}t_s(t)}{^{L}{}d_u(t)}\, {^{L}{}\dot d_u(t)} \, .
    \label{equation:Fg_lower}
\end{equation}
This can compensate for the elastic motion and minimize the forceps tip movement, i.e., keeping $^{T}{}d_p = 0$ by setting ${^{R}{}d_l(\tau) = - {}^{T}t_s(\tau)}$ with $\int^{t}_{0} \, {^{R}{}\dot d_l(\tau)} \, d\tau = -\int^{t}_{0} \frac{^{T}{}t_s(\tau)}{^{L}{}d_u(\tau)}\, {}^{L} \dot d_u(\tau)\, d\tau$.

Consequently, the input for the upper driver in the tissue grasping process with controlled $F_g$ can be set as 
\begin{equation}
    ^{L} \dot d_u(t) = f(\theta) (K_{P,1}{\,}e_1(t) + K_{I,1}\int_0^t {e_1(\tau){\,}{d \tau}} + K_{D,1} {\,} \dot e_1 (t)) \, ,
\label{Equation:Mode-I}
\end{equation}
where $e_1(t) = F^{\ast}_g(t) - F^e_g(t)$, $F^{\ast}_g(t)$ is the target grasping force, and $F^e_g(t)$ is the current estimated grasping force.

\begin{figure}[t!] 	 	 	
	\centering  	 	
	\includegraphics[width=0.488\textwidth]{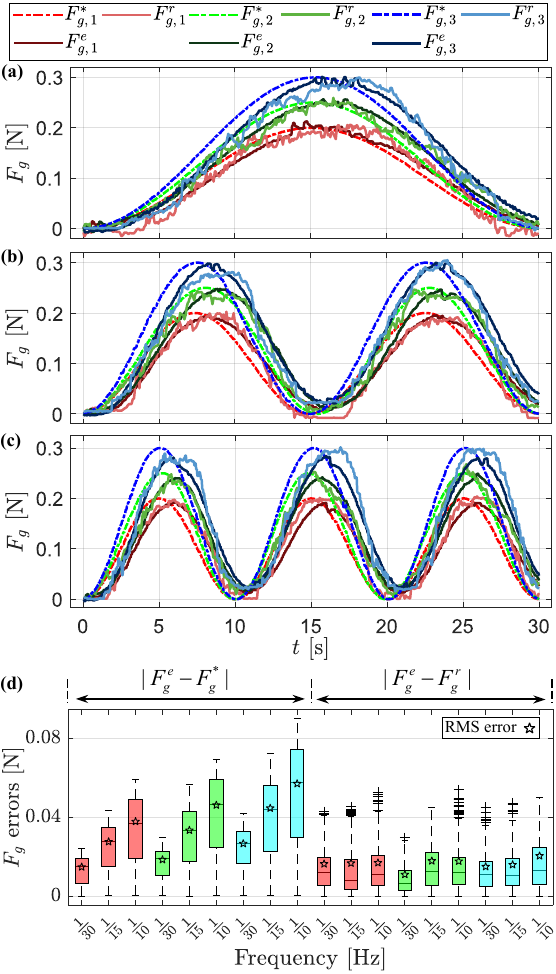}  	 	
	\caption{Evaluation results of tissue grasping with controlled $F_g$.
		(a), (b), and (c) are results of experiments conducted with $F^{\ast}_g$ frequency was set to {1/30Hz}, {1/15Hz}, and {1/10Hz}, respectively.
		$F^{\ast}_{g,\:,i}$, $F^e_{g,\:i}$, and $F^r_{g,\:i}$ indicate targeted, estimated, and reference grasping forces.
		$i=$ 1, 2, and 3 indicate experiments conducted with grasping angle $\theta = 10^\circ $, $30^\circ$, $50^\circ$, and the corresponding amplitudes of $F^{\ast}_{g,\: i}$ are 0.2N, 0.25N, and 0.3N.
        (d) Corresponding errors patched with red ($\theta=10^\circ$), green ($\theta=30^\circ$), and blue ($\theta=50^\circ$).
        The first nine boxes are the errors between $F^e_g$ and $F^\ast_g$, while the last nine are the errors between $F^e_g$ and $F^r_g$.
	} 	 	
	\label{Figure:Fg_data} 
\end{figure}

\subsubsection{Experimental Evaluation}
Fig.~\ref{Figure:Force_setup} shows the experimental setup, where a commercial pressure sensor (RFP603 200g, customized by RFP$^\circledR$, China) was adopted as the reference.
Associating with a linear voltage converter (K-CUT, China), the pressure sensor had a resolution of 0.01N in its linear sensing range [0, 0.6]N.
The forceps grasped the commercial pressure sensor, which directly measured the reference grasping force. 
Because the pressure sensor’s sensing area was a 10mm-diameter circle, 3D-printed jaws were rigidly fixed to the forceps’ jaws to ensure sufficient contact.
These extension jaws were configured with $\theta = 10^\circ $, $30^\circ$, and $50^\circ$ to provide different grasping angles.
Consequently, $l_3$ in (\ref{equation:grasping_force}) was set as $l^{\prime}_3$ in the following experiments.
To avoid the slide between the forceps and the pressure force sensor, we fixed the sensor to the lower extension jaw with a thin glue pad.
We tuned the controller parameters empirically by referring to the Ziegler–Nichols method. 
We observed acceptable performance with $[K_{P,\: 1},\: K_{I,\: 1}, \: K_{D,\: 1}] = [20, \, 1, \, 1]$ and applied them to the following evaluation experiments.

Because the grasping force is positively related to the grasping angle $\theta$, as shown in (\ref{equation:grasping_force}), the maximum grasping force that the forceps can produce varies with different $\theta$.
Moreover, to cover the range of $F_g$ estimation as much as possible, we set the amplitude of the target grasping force profile $F^{\ast}_{g}$ to 0.2N, 0.25N, and 0.3N for experiments with $\theta = 10^\circ $, $30^\circ$, and $50^\circ$, respectively, and note them by $i\in(1,2,3)$.
Referring to \cite{khadem2016modular,kim2015force,waters2022incipient,chua2022characterization}, we set the frequency of $F^{\ast}_{g}$ to {1/30}Hz, {1/15}Hz, and {1/10}Hz and repeatedly applied them to experiments with different $\theta$ configurations.

Fig.~\ref{Figure:Fg_data}(a), (b), and (c) present the grasping force results of experiments conducted with the targeted grasping force frequency equalling {1/30}Hz, {1/15}Hz, and {1/10}Hz, respectively.
For clarification, we plotted results with the same $F^{\ast}_g$ frequency in the same subplot.
$F_{g,\:i}^{\ast}$, $F^e_{g,\:i}$, and $F^r_{g,\:i}$ denote the input target grasping force, the forceps-estimated grasping force, and the measured reference grasping force, respectively.
Fig.~\ref{Figure:Fg_data}(d) shows the errors between $F^\ast_g$, $F^e_g$, and $F^r_g$ with boxplots, and the root mean square (RMS) errors are marked with pentagrams.
We also noticed that the error between $F^\ast_g$ and $F^e_g$ tended to increase when the frequency of the targeted force $F^\ast_g$ increased, and the maximum error (0.095N) was observed in the experiment with {1/10}Hz and $\theta = 50^\circ$.
On the contrary, errors between $F^\ast_g$ and $F^e_g$ showed slight variations, yet all were below 0.06N.
This indicates the grasping force $F_g$ can be estimated and controlled effectively during the grasping process.

\subsection{Tissue Pulling with Controlled $F_p$}
\label{section:pulling_control}
\subsubsection{Controller Design}
In our forceps, the current $F^e_p$ is also estimated and observed in each running cycle via (\ref{equation:fp}).
We assume the tissue has been grasped stably for the pulling stage, and the upper driver is locked by setting
\begin{equation}
    ^{L} \dot d_u(t) = 0.
\label{equation:Fp_upper}
\end{equation}
Then, the pulling force control can be achieved by directly relying on the lower driver’s motion.
Consequently, the input for the lower driver in the pulling process with controlled $F_p$ can be set as
\begin{equation}
    ^{R} \dot d_l(t) = K_{P,2}{\,}e_2(t) + {K_{I,2}} \int_0^t {e_2(\tau){\,}{d \tau}} + K_{D,2} {\,} \dot e_2(t) \, ,
\label{Equation:Mode-II}
\end{equation}
where $e_2(t) = F^{\ast}_p(t) - F^e_p(t)$, $F^{\ast}_p(t)$ is the target pulling force, and $F^e_p(t)$ is the current estimated pulling force.

\subsubsection{Experimental Evaluation}
Similarly, we evaluate the pulling force control by actuating the forceps to follow a series of inputted force profiles $F^\ast_p$ with different amplitudes and frequencies.
The experimental setup is shown in Fig.~\ref{Figure:Force_setup}, where the reference pulling force $F^r_p$ is measured by a commercial gauge force sensor (SBT630B 1kg, SIMBATOUCH, China) that is connected to the pressure sensor by an elastic. 
The forceps initially grasped the commercial pressure sensor with a grasping force $F_g = 0.2N$.
Then, the forceps pulled the pressure sensor with an input target pulling force profile $F^{\ast}_{p}$, having an amplitude of 2.0N.
$[K_{P,\: 2},\: K_{I,\: 2}, \: K_{D,\: 2}] = [10, \, 2, \, 5]$ were chosen and applied to the following evaluation experiments.
Similar to the tissue grasping evaluation, three frequencies, 1/30Hz, 1/15Hz, and 1/10Hz, were chosen and repeatedly applied to experiments with three grasping angles $\theta = 10^\circ $, $30^\circ$, and $50^\circ$ noted by $i\in(1,2,3)$.

\begin{figure}[t!] 	
	\centering 
	\includegraphics[width=0.488\textwidth]{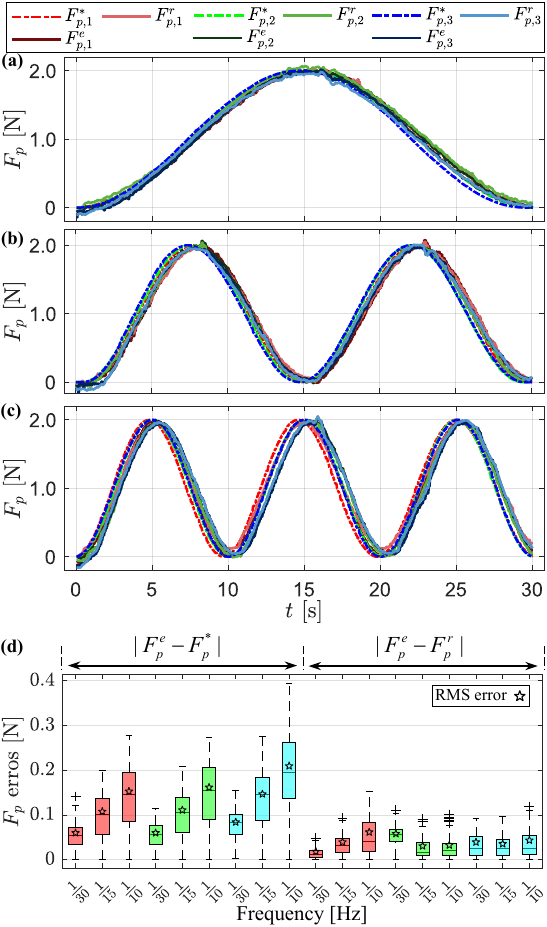} 
    	\caption{Evaluation results of the tissue pulling with controlled $F_p$. 
		(a), (b), and (c) present results of experiments conducted with $F^{\ast}_p$ frequency was set to 1/30Hz, 1/15Hz, and 1/10Hz, respectively, and the amplitude was 2.0N.
        $F^{\ast}_{p}$ and $F^{\ast}_{g,\:i}$ denote targeted pulling and grasping forces.
        $F^e_{p,\:i}$ and $F^e_{g,\:i}$ are the estimated forces by the forceps, while $F^r_{p,\:i}$ and $F^r_{g,\:i}$ are reference forces measured by commercial sensors.
        $i\in(1,2,3)$ denotes experiment conducted with grasping angle $\theta = 10^\circ $, $30^\circ$, and $50^\circ$.
        (d) Corresponding errors patched with red ($\theta=10^\circ$), green ($\theta=30^\circ$), and blue ($\theta=50^\circ$).
        The first nine boxes are the errors between $F^e_p$ and $F^\ast_p$, while the last nine are those between $F^e_p$ and $F^r_p$.}
	\label{Figure:Fp_data}
\end{figure}

\begin{figure}[t!] 	 	 	
	\centering  	 	
	\includegraphics[width=0.488\textwidth]{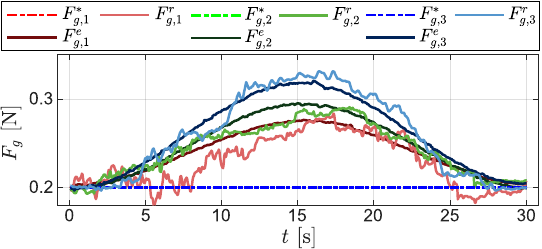}  	 	
	\caption{Grasping force $F_g$ variation resulting from coupling with pulling force $F_p$.
		$F^{\ast}_{g,\:,i}$, $F^e_{g,\:i}$, and $F^r_{g,\:i}$ indicate targeted, estimated, and reference grasping forces.
		$i=$ 1, 2, and 3, indicates experiments conducted with grasping angle $\theta = 10^\circ $, $30^\circ$, $50^\circ$.
        In these experiments, the amplitudes and frequency of $F^{\ast}_{p}$ were 2.0N and 1/30Hz.
	} 	 	
	\label{Figure:Fg_coupling} 
\end{figure}

Fig.~\ref{Figure:Fp_data}(a), (b), and (c) present the comparisons of pulling force $F_{p}$ in experiments conducted with $F^{\ast}_{p}$ frequency equal to {1/30}Hz, {1/15}Hz, and {1/10}Hz, respectively.
For clarification, we plotted results with the same $F^{\ast}_p$ frequency in the same subplot.
$F_{p,\:i}^{\ast}$, $F^e_{p,\:i}$, and $F^r_{p,\:i}$ denote the input target pulling force, the forceps' estimated pulling force, and the measured reference pulling force, respectively
Fig.~\ref{Figure:Fp_data}(d) shows the errors between $F^\ast_p$, $F^e_p$, and $F^r_p$ with boxplots, and RMS errors are marked with pentagrams.
The error between $F^\ast_p$ and $F^e_p$ tended to increase when the frequency of $F^\ast_p$ increased, and the maximum error 0.4N was observed in the experiment with {1/10}Hz and $\theta = 50^\circ$.
On the contrary, errors between $F^e_p$ and $F^r_p$ showed slight variations, yet all were below 0.16N.
This indicates that $F_p$ can be estimated and controlled efficiently during the pulling process.


\subsection{Tissue Traction with Controlled $F_g$ and $F_p$}
\subsubsection{Integrated Controller for Grasping and Pulling Stages}
The results above have shown the potential of the proposed controllers in automatic tissue traction, which consists of a grasping stage followed by a pulling stage. 
In particular, the grasping stage can directly employ the controller developed in Sec. \ref{section:grasping_control}, i.e., pure grasping force control using (\ref{equation:Fg_lower}) and (\ref{Equation:Mode-I}). 

The pulling stage may also just use the pure pulling force controller proposed in Sec. \ref{section:pulling_control}; however, a notable coupling between $F_p$ and $F_g$ was observed in the experiments in Sec. \ref{section:pulling_control}.
For example, Fig.~\ref{Figure:Fg_coupling} presents the $F_g$ variation resulting from the changed $F_p$ corresponding to experiments in Fig.~\ref{Figure:Fp_data}(a), and the maximum values of these variations were 0.073N, 0.096N, and 0.116N for $\theta = 10^\circ,\, 30^\circ,$ and $50^\circ$, respectively. 
This force coupling phenomenon was also observed in \cite{kim2018sensorized, seok2019compensation, liu2023hapticsenabled} when these studies evaluated their forceps' force-sensing capabilities.

The force coupling phenomenon can be explained by (\ref{equation:fp}) and (\ref{equation:grasping_force}).
Referring to them, we can formulate grasping force as
\begin{equation}
	F_g =  \frac{F_d \, l_2 \sin{\alpha}}{2l_3} = \frac{(F_p+F_s) \, l_2 \sin{\alpha}}{2l_3} \, ,
	\label{equation:force_coulping}	
\end{equation}
where $F_d = F_p+F_s$.
This indicates that the grasping force $F_g$ is positively related to the pulling force $F_p$.

When dealing with delicate and critical tissues, it is often desirable to maintain a controlled grasping force while pulling the tissue.
Therefore, we propose to simultaneously control the grasping and pulling forces during the tissue pulling stage.
This can be conveniently achieved by setting the inputs for lower and upper drivers in the pulling process with (\ref{Equation:Mode-II}) and (\ref{Equation:Mode-I}), respectively.

\subsubsection{Experimental Evaluation}
Here, we adopted the same experimental setup shown in Fig.~\ref{Figure:Force_setup}.
The experimental procedures simulated the tissue traction process with the following considerations.
First, the forceps grasp the pressure sensor following a target grasping force profile $F_g^{\ast}$.
Then, during the pulling process, the forceps pull the pressure sensor following a target pulling force profile $F^{\ast}_p$ while keeping the grasping force at a constant value $F^{\ast, \,t}_g$. 
$F^{\ast, \,t}_g$ simulated a threshold to limit the grasping force and avoid unexpected tissue breaking.

\begin{figure}[t!] 	 	 	
	\centering  	 	
	\includegraphics[width=0.488\textwidth]{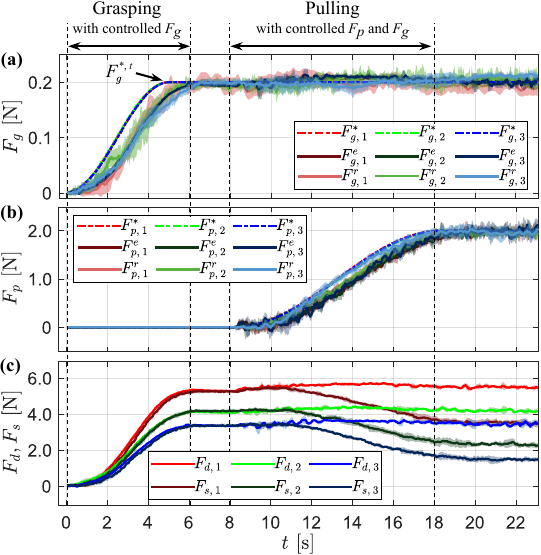}   	
	\caption{Evaluation results of tissue traction with controlled $F_g$ and $F_p$.
     (a) and (b) Comparisons of grasping force $F_g$ and pulling force $F_p$ results. 
     $F^r_{g,\:i}$ and $F^r_{p,\:i}$ are measured reference forces, $F^e_{g,\:i}$ and $F^e_{p,\:i}$ indicate the estimated forces, $F^{\ast}_{g}$ and $F^{\ast}_{p,\:i}$ are targeted forces, and $F^{\ast,\: t}_g$ (0.2N for all experiments) is the threshold target force for grasping.
     $i\in(1,2,3)$ denotes experiment conducted with grasping angle $\theta = 10^\circ $, $30^\circ$, and $50^\circ$. 
     (c) Variations of driving force $F_d$ and spring elastic force $F_s$.} 	 	
	\label{Figure:FpFg_data} 
\end{figure} 

\begin{table}[t!] 	 	 	
    \centering 	 	 	
    \caption{Worst-case Errors in The Pulling Phase of Simultaneous Multiple Force Control}  
    \begin{tabular}{ccccccc} 		 		 		
    \toprule 		
    Force & Comparison & Error & $10^\circ$ & $30^\circ$& $50^\circ$ \\ \toprule 	
    \multirow{6}{*}{$F_g$ [N]}&                       & Mean& 0.0054 & 0.0071 & 0.0106 \\
                              & ($F^\ast_g,\, F^e_g$) & RMS & 0.0063 & 0.0083 & 0.0128 \\ 
                              &                       & Max & 0.0151 & 0.0202 & 0.0377 \\ 
    \cline{2-6} 	
    \specialrule{0em}{1pt}{1pt}
                              &                       & Mean& 0.0144 & 0.0140 & 0.0171 \\
                              & ($F^e_g,\, F^r_g$)    & RMS & 0.0163 & 0.0167 & 0.0201 \\ 
                              &                       & Max & 0.0345 & 0.0448 & 0.0532 \\ 
    \hline	
    \specialrule{0em}{1pt}{1pt}                          
    \multirow{6}{*}{$F_p$ [N]}&                       & Mean& 0.2480 & 0.2024 & 0.2019 \\
                              & ($F^\ast_p,\, F^e_p$) & RMS & 0.2890 & 0.2457 & 0.2385 \\ 
                              &                       & Max & 0.5529 & 0.5382 & 0.5218 \\ 
    \cline{2-6} 	
    \specialrule{0em}{1pt}{1pt}
                              &                       & Mean& 0.1043 & 0.1122 & 0.1920 \\
                              & ($F^e_p,\, F^r_p$)    & RMS & 0.1226 & 0.1331 & 0.2253 \\ 
                              &                       & Max & 0.3068 & 0.3374 & 0.4437 \\ 
    \toprule 
    \end{tabular} 	 
    \label{Table:FpFg_control}   
\end{table}	

Specifically, the targeted grasping force $F^{\ast}_g$ first gradually increased from zero to the threshold $F^{\ast,\,t}_g = 0.2$N during the grasping phase ($0-5$s).
Then, it kept constant during the pulling phase ($8-18$s) when the target pulling force $F^{\ast}_p$ increased from 0 to 2.0N.
Finally, both $F^{\ast}_g$ and $F^{\ast}_p$ kept constant during ($18-25$s).
During ($5-8$s), $F^{\ast}_g$ and $F^{\ast}_p$ were constant to visually separate the grasping and pulling phases.

\begin{figure*}[t!] 	  		
    \centering  	 		
    \includegraphics[width=1.0\textwidth]{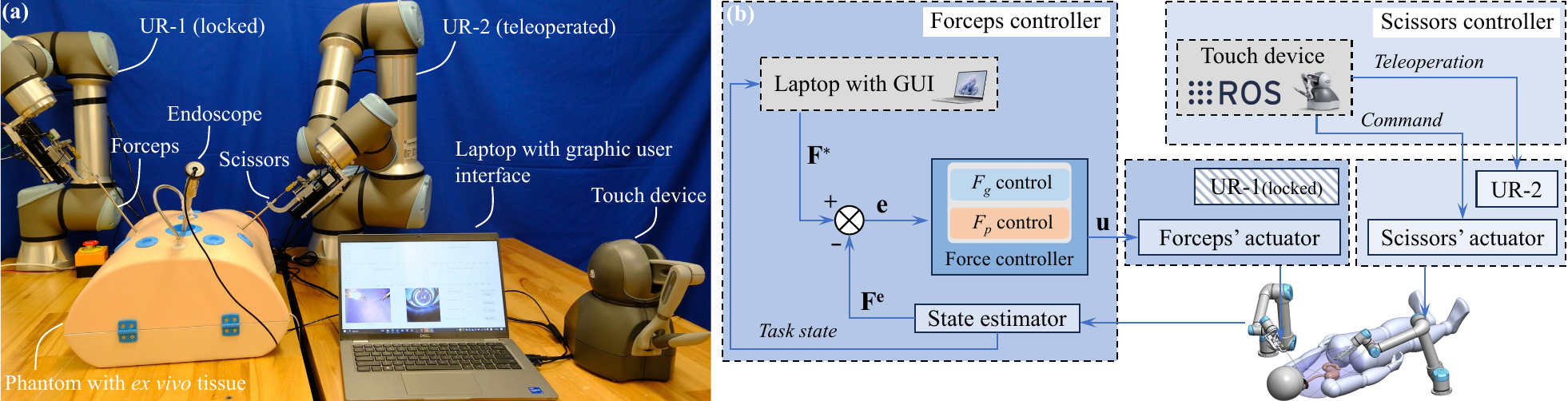} 
    \caption{(a) Experimetal setup of robotic \textit{ex vivo} tissue resections with automatic traction.
    A human body phantom was used for placing \textit{ex vivo} chicken tissue simulating lesions.  
    The micro-level actuators with instruments were mounted on two universal robotic arms, UR-1 and UR-2.
    Specifically, UR-1, carrying the forceps tissue grasping and pulling, was hocked to the initial pose and locked during the tissue resection procedure, while UR-2 was teleoperated by a touch device to adjust the pose of the scissors for tissue cutting.
    The instruments and endoscopy were implemented through simulated minimally invasive surgical ports on the phantom.
    The laptop provided the operator with a graphic user interface (GUI) to monitor the procedure and input orders.
    (b) Integrated controller diagram of the dual-arm robotic system.
    } 
    \label{Figure:robotic_setup}     		
\end{figure*}

Each experimental procedure was repeated five times.
The standard deviation plots Fig.~\ref{Figure:FpFg_data}(a) and (b) show the comparison of estimated, targeted, and reference values of $F_g$ and $F_p$, respectively.
Since all the experiments followed the same $F^\ast_g$ and $F^\ast_p$ profiles, their results almost overlaid.
In these traction experiments, we are more concerned with the force variations during the pulling process, and Table~\ref{Table:FpFg_control} lists the worst-case errors (see Appendix-A) of $F_g$ and $F_p$ during this period.

\begin{figure*}[t!] 	 	 
		\centering  	 	
		\includegraphics[width=1.0\textwidth]{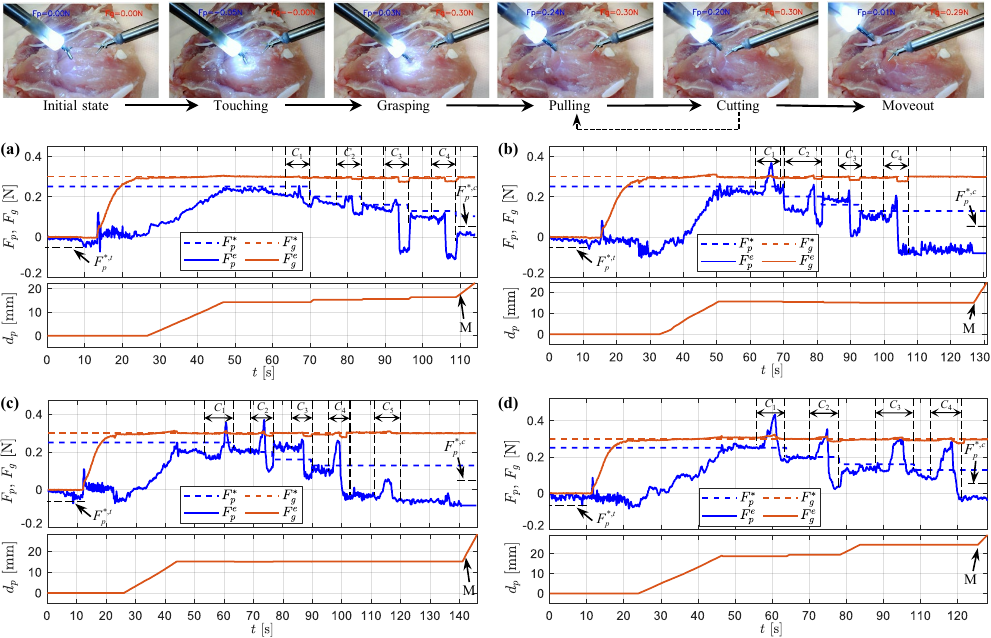}	
		\caption{
        Results of automatic traction-assisted tissue resections on \textit{ex vivo} chicken tissue, where the top endoscopy snapshots show the critical stages of the operation flow.
        (a), (b), (c), and (d) present the results of four procedures, where $F_g$, $F_p$, and $d_p$ denote the grasping force, pulling force, and pulling distance, respectively.
        Four procedures followed the same operation flow, wherein the forceps performed automatically, while the scissors relied on teleoperation to adjust each cut’s pose.
        The cuts performed in each resection are indicated by $C_i$, and the moving out is denoted by M.  
        } 	 	
	\label{Figure:exvivo}  
\end{figure*} 

We can see the mean, RMS, and maximum errors between $F^\ast_g$ and $F^e_g$ are all under (0.0106, 0.0128, 0.0377)N, which were comparable with that of experiments shown in Fig.~\ref{Figure:Fg_data}.
The mean, RMS, and maximum errors between $F^\ast_p$ and $F^e_p$ are all under (0.2480, 0.2890, 0.5529)N, which were also comparable to that in experiments shown in Fig.~\ref{Figure:Fp_data}.
Moreover, compared to that without controlled grasping force (see Fig.~\ref{Figure:Fg_coupling}), the maximum deviation of $F_g$ in the pulling phase reduced by around 300\% for $\theta = 50^\circ$ (from 0.116N to 0.0377N), 470\% for $\theta = 30^\circ$ (from 0.096N to 0.0202N), 460\% for $\theta = 10^\circ$ (from 0.073N to 0.0151N).
These indicate that simultaneous $F_p$ and $F_g$ control can be effectively achieved during the pulling process.

In addition, Fig.~\ref{Figure:FpFg_data}(c) shows the corresponding driving force $F_d$ and spring elastic force $F_s$ of these experiments.
We can see that less $F_d$ and $F_s$ were required for forceps with larger $\theta$ to achieve the same $F^{\ast,\,t}_g$.
Moreover, the driving force $F_d$ remained nearly constant during the pulling phase while the spring elastic force $F_s$ (equal to the supporting force) gradually decreased.
This was because the constant grasping force required constant driving force, and the system had to reduce the elastic force $F_s$ to ensure the increased pulling force $F_p$.
This phenomenon confirmed the analysis in (\ref{equation:force_coulping}) and revealed that the decoupling control of $F_g$ and $F_p$ was achieved by indirectly controlling $F_d$ and $F_s$.


\section{Automatic Traction-Assisted Tissue Resection}
\label{Section:ex_vivo_experiment}

Tissue traction is frequently performed in surgical operations such as lesion sampling and tumor dissection.
Here, we demonstrate the feasibility of the proposed method for automatic tissue traction in a dual-arm tissue resection task on \textit{ex~vivo} tissues.

\subsection{Experimental Setup and Operation Flow}
\subsubsection{Experimental Setup}
Fig.~\ref{Figure:robotic_setup}(a) shows the experimental setup, where two UR robotic arms were adopted as the macro-level actuators.
The \textit{ex~vivo} chicken tissue was placed in a human body phantom to simulate the lesion, and instruments were implemented through simulated minimally invasive ports on the phantom.
Both universal robotic arms, UR-1 and UR-2, have been pre-docked to the initial state, and the micro-level actuators were responsible for performing the automatic traction-assisted tissue resection task locally.
During the tissue traction, no manual manipulation was applied to the forceps.
Specifically, UR-1, carrying the forceps for tissue traction, was locked during the operation, while UR-2 was teleoperated to adjust the pose of the scissors for tissue cutting.

To integrate the force controller and the robotic controller, we further designed a software architecture, as depicted in Fig.~\ref{Figure:robotic_setup}(b), which contains a forceps controller and a scissors controller.
The forceps controller was responsible for the tissue traction.
Dell Latitude 5420 laptop equipped with an i5-11${th}$ @ 2.6GHz CPU received the task state and sent the target forces $\mathbf{F}^\ast$ to the force controller.
Subsequently, the force controller generated the inputs $\mathbf{u} = [^{R}\dot d_l,\,  ^{L}\dot d_u ]^\top$ and sent them to the forceps’ micro-level actuator of the forceps. 
The operator could supervise the procedure and input orders through a designed graphic user interface (GUI), shown in Fig.~\ref{Figure:robotic_setup}(a).
This GUI also displayed real-time data, including sensed multiple forces, pulling distance, endoscope image, and camera frame of the sensing module.

The scissors controller was responsible for the tissue cutting.
A touch device (3D Systems, USA), shown in Fig.~\ref{Figure:robotic_setup}(a), enabled the teleoperation of the UR-2 robotic arm and facilitated the transmission of cutting commands to the micro-level actuator of the scissors.
User movements captured by the touch device were abstracted by the touch ROS driver provided by \cite{mathur2019evaluation}. 
These movements were then transformed and relayed to the robot's end-effector frame, located at the tip of the scissors.

The force controller, state estimator, and signal generator for micro-level actuators were implemented on an STM32H7 microcontroller, which communicated with the laptop and ROS via serial ports.
The UR-2 robot was connected to another computer running ROS via an Ethernet cable, and the touch device was linked to the computer through a USB port. 
The publishing rate for the UR-2 robot was 500 Hz, while the touch device’s was 1000 Hz. 
Each pose trajectory command sent to the UR-2 robot had a duration of 1 ms.

\subsubsection{Resection with Automatic Tissue Traction}

For typical tissue resection, we noticed that the required pulling force at the first cut is more significant than that of the following cuts.
Therefore, the targeted pulling force $F_p^{\ast}$ should be reduced after each cut.
Moreover, because of the elasticity, the tissue body tends to collapse after the connection part has been cut out.
This indicates that the cutting surface may passively be revealed by the tissue collapse when the tissue has been pulled up to a certain displacement.

Considering the above phenomena and patterning after the clinical dissection procedure \cite{skandalakis2021surgical}, we designed an operation flow with automatic traction.
The automatic traction was built upon predefined grasping and pulling forces profiles, complemented by the pulling distance information.
The forceps first reached the tissue with touch detection by comparing the estimated $F_p$ with a set threshold $F_p^{*,\:t}=-0.05$N (the negative sign indicates the operation is opposite to pulling tissues).
Next, the tissue was grasped to the target grasping force $F_g^{\ast} = 0.3$N, followed by pulling up and holding with an initial target pulling force $F_p^{\ast,\: 0} = 0.25$N.
After the pulling, the scissors were teleoperated to the target tissue, and cutting was performed. 
Then, the operation alternated between pulling and cutting several times, and the target pulling force $F_p^{\ast,\: i}$ reduced to $\rho F_p^{\ast,\: i-1}$ after the $i^{th}$ cut.
Here, $i \in[1,\:m]$, $m$ is the number of cuts performed, and $\rho = 0.8$ is the reduction ratio.
Once reaching the target force $F_p^{\ast,\: i}$ in each pulling, the forceps kept static with no pulling distance $d_p$ change to provide a constant cutting surface.
The incremental pulling distance $\Delta d_p$ between cuts was limited by a threshold $\Delta d_p^t = 20$mm, and the total pulling distance $d_p$ was limited by $d_p^t = 30$mm.
A threshold for cutoff detection $F^{\ast,\,c}_p = 0.05$N was also included.
When the estimated pulling force $F^e_p$ after each cut was less than $F^{\ast,\,c}_p$, the forceps would not pull up the tissue further.
Then, the operator was required to check whether the tissue had been cut off and confirm whether another cut was needed or could be moved out directly.
The above operation flow is also summarized in Appendix-B. 

\subsection{Tissue Resection Results}
Here, the automatic tissue grasping process was achieved by (\ref{equation:Fg_lower}) and (\ref{Equation:Mode-I}), and the pulling process relied on (\ref{Equation:Mode-II}) and (\ref{Equation:Mode-I}).
We conducted four resections, and their results are shown in Fig.~\ref{Figure:exvivo}(a), (b), (c), and (d), including grasping force $F_g$, pulling force $F_p$, and pulling distance $d_p$.
The critical snapshots of the experimental procedure are also presented at the top of Fig.~\ref{Figure:exvivo}.

Four procedures were carried out with the same piece of tissue, but the localized conditions of the grasped tissues differed, and the scissors' pose was teleoperated.
As a result, the number of cuts ($m$) performed and the operation time ($t$) varied among these procedures. 
Specifically, for the experiments presented in Fig.~\ref{Figure:exvivo}(a), (b), and (d), four cuts were performed ($m=4$), while for the experiment in Fig.~\ref{Figure:exvivo}(c), five cuts were performed ($m=5$), as indicated by $C_i$.
The rapid variations in pulling force $F_p$ resulted from interactions between scissors and tissue.
When the forceps reach the cutting surface to perform the tissue cutting, they may stretch or press the cutoff tissue, which can cause the increase or decrease of pulling force $F_p$ sensed by the forceps.
At this stage, the sensed pulling force $F_p$ is the stretch force applied on the cutoff tissue between the forceps and scissors.
Although the sensed $F_p$ sometimes exceeded the target $F^\ast_p$, the tissue was not subjected to excessive pulling or pushing, thereby not compromising patient safety, as the forceps kept static during each cutting phase.
All procedures successfully cut the tissue off, indicating that automatic tissue traction with multi-force control was effectively implemented and had the potential to promote autonomous MIS.

\section{Discussion}
\label{Section:discussion}

\subsection{Force Sensing and Tracking}
\subsubsection{Sensing Resolution and Range}
Although through the distributed software framework design and direct memory access (DMA) communication, the local controller could achieve high-frequency operation with minimized signal interference, the updating frequency of estimated elastic supporting force $F_s$ was limited to 30Hz by the camera (OVM6946, Omnivision Inc., USA) update rate.
The frequency could be improved by adopting a camera with a higher rate.
Consequently, the performance of force estimation and control could be further improved.
Replacing the spring with flexures that have different stiffness, the sensitivity and resolution of the sensing module can also be configured.
Additionally, the grasping force $F_g$ estimation is related to the supporting force $F_s$, and the sensible range of $F_s$ (limited by the spring’s deformation) is $[0-5]$N in our prototype.
Therefore, the sensible range of $F_g$ is also limited and positively related to grasping angle $\theta$.

\begin{figure}[t!] 	 	 	
	\centering  	 	
	\includegraphics[width=0.488\textwidth]{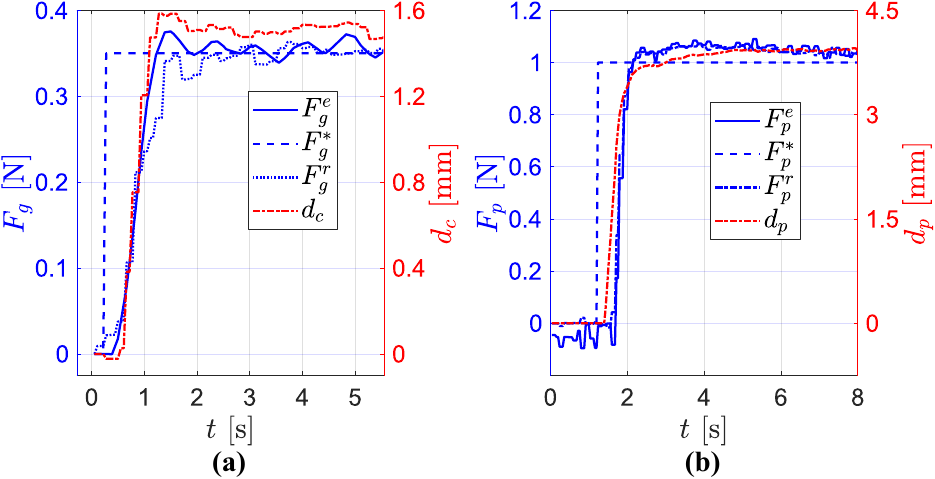}   	
	\caption{Force tracking performance of step signal input.
		(a) $F^e_{g}$, $F^\ast_{g}$, and $F^r_{g}$ are estimated, target, and reference grasping forces. $d_c$ is the displacement of the driving cable. 
        (b) $F^e_{p}$, $F^\ast_{p}$, and $F^r_{p}$ are estimated, target, and reference pulling forces. $d_p$ is the pulling distance.}	 	
	\label{Figure:FpFg_step} 
\end{figure}

\begin{figure}[t!] 
	\centering  	
	\includegraphics[width=0.488\textwidth]{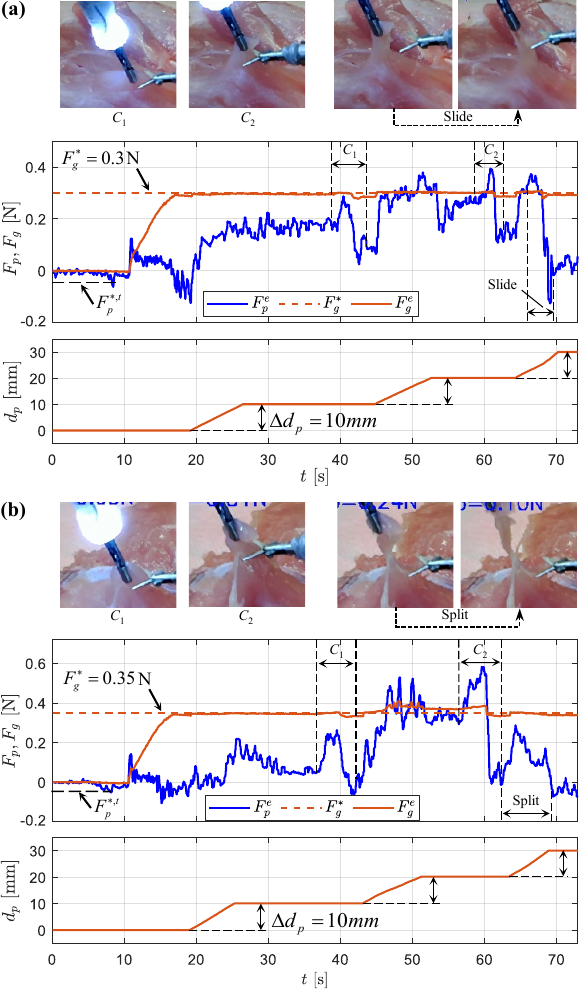} 
		\caption{Tissue resection on chicken tissue without pulling force $F_p$ control. (a) and (b) show tissue slide and split due to excessive pulling forces, respectively. $\Delta d_p$ indicates the incremental pulling distance between each two cuts. $F_g$ and $F_p$ denote the recorded grasping force and pulling force. Tissue cutting is indicated by $C_i$.}	
	\label{Figure:exvivo_dp} 
\end{figure}

\subsubsection{Force Tracking}
For the PID controller, the proportional term produces an output value proportional to the current error value and generally operates with a steady-state error.
Then, the integral term is required to eliminate the steady-state error.
However, a large proportional or integral term could result in a notable overshoot \cite{ang2005pid}.
This overshoot can introduce excessive forces and bring risks to tissue manipulation. 
Consequently, we tended to select smaller $K_P$ and $K_I$ values during the controller tuning process.
In addition, a large derivative term may impact the controller's stability, but a small one can improve it \cite{ang2005pid}.
Therefore, smaller $K_D$ values were also preferred.
These parameters can be fine-tuned to enhance the performance in specific aspects for different preferences.

We demonstrated the system has the ability to track force profiles with different frequencies and amplitude, as shown in Fig.~\ref{Figure:Fg_data}, Fig.~\ref{Figure:Fp_data},  and Fig.~\ref{Figure:FpFg_data}.
Because the actuators’ displacements and speeds can not change rapidly, the response speed of force control is limited, and there are rise times for the system to track the target forces.
These rise times approximately equal the execution times of the grasping and pulling tasks.
For example, using the experimental setup shown in Fig.~\ref{Figure:Force_setup}, we verified the system's step response, and the results are shown in Fig.~\ref{Figure:FpFg_step}.
For target grasping force $F_g^\ast$ with 0.35N step signal, a 1.5s rise time with 1.6mm driving cable displacement $d_c$ was observed.
For target pulling force $F_p^\ast$ with a 1.0N step signal, a 0.5s rise time with 4mm pulling distance $d_p$ was observed.
However, the force variation in tissue traction often involves considerable tissue deformation, and no rapidly changed force is required \cite{skandalakis2021surgical}.
In addition, \textit{ex vivo} experiments show that the forceps can effectively complete the adjustment of forces between cuts. 
Therefore, these rise times will not compromise the effectiveness of tissue traction.

\subsection{Automatic Traction}

The tissue traction strategy proposed in Section~\ref{Section:ex_vivo_experiment} seems straightforward. 
However, we still observed some inspiring phenomena by comparing and analyzing these results.
From Fig.~\ref{Figure:exvivo}, we can see that for the same target pulling force $F^{\ast,\,i}_p$, the required pulling distance $d_p$ varied.
Specifically, $d_p$ was almost constant in Fig.~\ref{Figure:exvivo}(b) and (c), while it gradually increased in Fig.~\ref{Figure:exvivo}(a) and (d).
This indicates that tissue pulling that solely relies on pulling distance $d_p$ has a higher potential for tissue split due to excessive pulling forces.
On the other hand, we must admit that tissue split cannot be avoided by simply relying on the pulling force information either.
This is because when tissue has been cut very finely, the pulling force exerted on the forceps will be too subtle to be reliably sensed and controlled.
That is why incremental and total pulling distances should be limited in tissue traction, and the operator was required to check the tissue condition when the pulling force $F^e_p$ was less than a threshold $F^{\ast, \, c}_p$ (0.05N in our experiments).
This tissue condition-checking task has the potential to be automated with more input information, for example, the fluorescence of tumor \cite{woo2021fluorescent}. 

\begin{figure}[t!] 	 	
	\centering  	
	\includegraphics[width=0.488\textwidth]{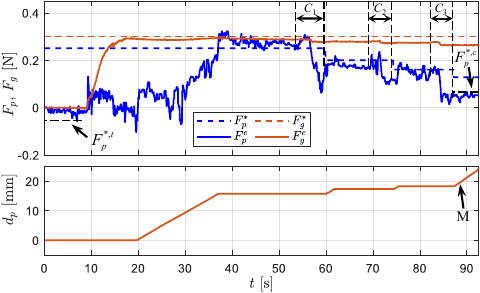} 
		\caption{Tissue traction on chicken tissue without decoupling force control, wherein the grasping force gradually decreased due to instrument interactions (between scissors and tissue). 
        $F_g$, $F_p$, and $d_p$ are grasping force, pulling force, and pulling distance results.
        The number of performed cuts is three, each cut is indicated by $C_i$, and the moving out by M.
        }	
	\label{Figure:exvivo_without} 
\end{figure}

\begin{figure}[t!] 	 	
	\centering  	
	\includegraphics[width=0.488\textwidth]{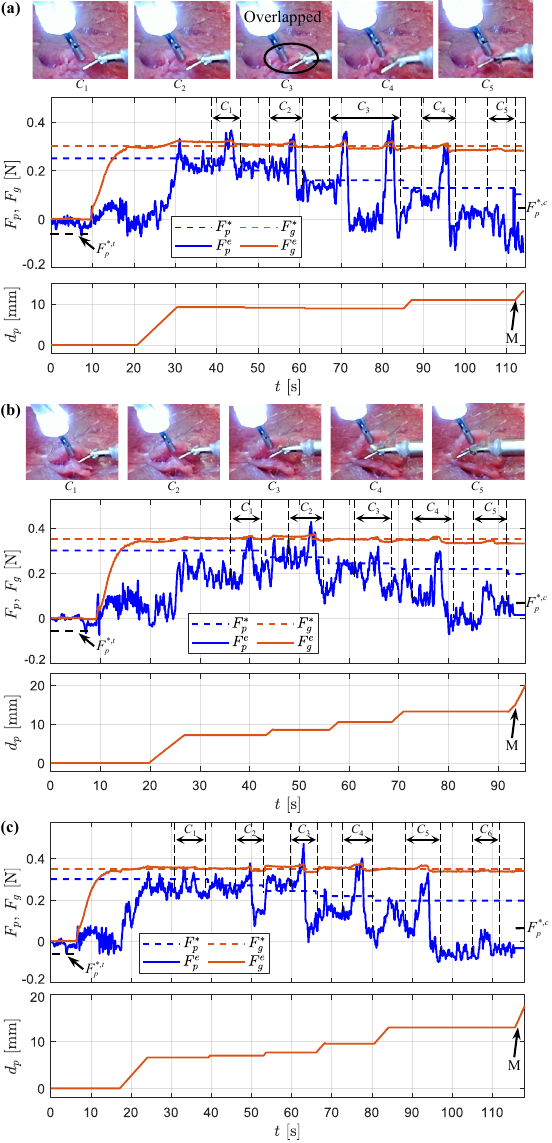} 
		\caption{Tissue resection on beef tissues. (a) Overlapped cutting surface due to insufficient pulling force $F_p$, where the parameters are the same as the experiments on chicken tissue in Fig.~\ref{Figure:exvivo}. (b) and (c) Improved cutting surface with increased pulling forces $F_p$. $F_g$, $F_p$, and $d_p$ denote the grasping force, pulling force, and pulling distance, respectively. Tissue cutting and moving out are indicated by $C_i$ and M.}
	\label{Figure:exvivo_beef} 
\end{figure}

\subsubsection{Traction without $F_p$ Control}
To further identify the potential benefits of tissue traction with controlled pulling force $F_p$, we conducted a group of resections on \textit{ex vivo} tissue relying on pulling distance $d_p$.
During the operation, the incremental pulling distance $\Delta d_p$ was set to 10mm, while the target pulling force $F^\ast_p$ and its reduction ratio $\rho$ were disabled.
Fig.~\ref{Figure:exvivo_dp}(a) and (b) show that the tissue slid and split due to excessive pulling forces $F^\ast_p$, respectively.
For each experiment, $F_g$, $F_p$, and $d_p$ were recorded, and the top endoscopy snapshots show the critical events of the operation.

In Fig.~\ref{Figure:exvivo_dp}(a), the target grasping force $F^\ast_g$ was set to 0.3N as that in Fig.~\ref{Figure:exvivo}.
Because the tissue was pulled up with $\Delta d_p$ after each cut without intentional $F_p$ control, the excessive $F_p$ was observed between 50$-$53s with $F_p$ = 0.38N (which was limited by a threshold of 0.25N in Fig.~\ref{Figure:exvivo}) and the tissue slide occurred at 66s as the tissue was excessively pulled with $d_p$ = 25 mm and $F_g$ = 0.37N.

We then increased the target grasping force $F^\ast_g$ to 0.35N in the experiment presented in Fig.~\ref{Figure:exvivo_dp}(b).
Although the tissue slide was avoided, we observed tissue split.
The excessive $F_p$ was observed between 45$-$55s with $F_p$ = 0.4N, and the tissue slide occurred at 63s as the tissue was excessively pulled with $d_p$ = 22mm and $F_p$ = 0.24N.
This tissue split could be avoided with controlled $F_p$, as the estimated $F_p$ after $C_2$ was 0.03N at 61s, which is less than the cutoff detection threshold $F^{\ast,\,c}_p = 0.05$N in Fig.~\ref{Figure:exvivo}.
This means the remaining tissue was too fine and should not be pulled up further after $C_2$.

\subsubsection{Traction without Decoupling Force Control}
We also conducted tissue traction with the grasping process achieved by (\ref{equation:Fg_lower}) and (\ref{Equation:Mode-I}), and the pulling process relied on  (\ref{Equation:Mode-II}) and (\ref{equation:Fp_upper}), i.e., the grasping force $F_g$ is not controlled during the pulling process. 
The tissue also has been successfully cut off, and the result is shown in Fig.~\ref{Figure:exvivo_without}.
Since the required $F_p$ in the tissue-pulling process is small (0.25N in these experiments), the force coupling was not obviously observed.
However, another interesting phenomenon caught our attention: the grasping force $F_g$ in Fig.~\ref{Figure:exvivo_without} gradually decreased (0.04N reduction at the end, 12\% of 0.3N) after each cut.
This could result from the interaction between the scissors and tissue, which caused rapid force variation and impaired the grasping condition.
This reduced grasping force could increase the risk of tissue slides in the following operations.

Nevertheless, with decoupling force control, the grasping force $F_g$ in the experiments presented in Fig.~\ref{Figure:exvivo} recovered to the target value $F_g^\ast$ (with 0.01N errors at the end, 3\% of 0.3N).
Compared to the experiment in Fig.~\ref{Figure:exvivo_without} that has a 12\% reduction of $F_g$, enabling grasping force control in the pulling process could reduce tissue slide risk.
This phenomenon further supports the criticality of simultaneous multi-force control in tissue manipulations.

\subsubsection{Automatic Traction on Different Tissues}
The used parameters in tissue resections, including touch detection threshold $F^{\ast,\,t}_p$, targeted grasping force $F_g^{\ast}$, initial targeted pulling force $F_p^{\ast}$, reduction ratio $\rho$, pulling distance incremental threshold $\Delta d_p^t$, total pulling distance threshold $d_p^t$, and cutoff detection threshold $F^{\ast,\,c}_p$, were empirically chosen.
For more precise operation, these parameters should be identified according to clinical requirements.
To help understand this, we also conducted automatic traction-assisted resections on \textit{ex vivo} beef tissue.

We first conducted a resection with the same parameters in Fig.~\ref{Figure:exvivo}, i.e., initial target pulling force $F^{\ast,0}_p$ = 0.25N, reduction ratio $\rho$ = 0.8, and target grasping force $F^\ast_g$ = 0.3N.
Fig.~\ref{Figure:exvivo_beef}(a) presents the result, and the top snapshots show the critical stages.
Although the tissue was successfully cut off with five cuts, the operation was more challenging as the cutting surface overlapped due to insufficient pulling force $F_p$, which made it more difficult for the scissor to reach, as shown in $C_3$ of the snapshots.
This may be because beef tissue is tougher than chicken tissue \cite{HuiNip2001BeefTexture}.

Consequently, we slightly increased $F^{\ast,0}_p$ to 0.3N, $\rho$ to 0.9, and $F^\ast_g$ to 0.35N to provide a better cutting surface.
Then, two resections were repeatedly conducted, and the results are shown in Fig.~\ref{Figure:exvivo_beef}(b) and (c).
The top snapshots are the critical stages corresponding to the trial of Fig.~\ref{Figure:exvivo_beef}(b).
The snapshots show that the cutting surfaces in Fig.~\ref{Figure:exvivo_beef}(b) have been improved compared to Fig.~\ref{Figure:exvivo_beef}(a), and no tissue slide, break, or split occurred.
	
In practical surgery, external sensing modules, such as a vision system, could also provide more information for decision-making.
With additional information and considerations of clinical constraints, the tissue resection strategy has the potential to be extended to more general operations.

\subsection{Other Applications and Future Work}
This paper demonstrated the potential application of multi-force control for automatic tissue traction by forceps through robotic \textit{ex vivo} resections with simulated surgical setup.
In the future, we will conduct more systematic verification with practical clinical setups under the guidance of clinicians.
The proposed force controller has the potential to be extended to other surgical operations as well.
These include but are not limited to robotic palpation and tissue handling compensating for external motion disturbances.
Considering the clinical constraints, these extended characteristics will be further studied in future work with other tissue manipulation tasks.
Comparing the surgical outcomes of automated and manual operations in terms of duration, incision, bleeding, etc., is also significant and will be explored in future work.

Because the grasping force and pulling force can be controlled separately or simultaneously, more alternatives are provided to the operator for different applications. 
Moreover, with the one-step-docking design, the instrument can be replaced and delivered quickly.
Specifically, the scissors and forceps were actuated by the same micro-level actuator in our prototype, as shown in Fig.~\ref{Figure:robotic_setup}(a).
This could also bring benefits to sterilization and interoperative instrument exchanges.

\section{Conclusion}
\label{Section:conclusion}
This paper has introduced a method for automating tissue traction with miniature forceps featuring controlled grasping and pulling forces.
Specifically, the grasping stage relied on a controlled grasping force, while the pulling stage was under the guidance of a controlled pulling force.
In particular, during the pulling process, the simultaneous control of both grasping and pulling forces was also enabled for more precise tissue traction, achieved through force decoupling.
The force controller was built upon a static model of tissue manipulation, considering the interaction between the miniature force-sensing forceps and soft tissue.
The efficacy of this force control approach was validated through a series of experiments comparing targeted, estimated, and actual reference forces. 
Various tissue resections were conducted on \textit{ex vivo} tissues employing a dual-arm robotic setup to verify the feasibility of the proposed method in surgical applications.
Finally, we discussed the benefits of multi-force control in tissue traction, evidenced through comparative analyses of various \textit{ex vivo} tissue resections with and without the proposed method, and the potential generalization with traction on different \textit{ex vivo} tissues.
The results affirmed the feasibility of automatic tissue traction using miniature forceps with multi-force control, suggesting its potential to promote autonomous MIS.
In the future, this approach will be extended to accommodate other tissue manipulations and take into account more strict clinical constraints and surgical versatility.

\section*{Appendix}
\label{Appendix}
\subsection{Worst-Case Error Calculation}
The worst-case errors in Table~\ref{Table:FpFg_control} were calculated by 
\begin{align*}
    \left\{
    \begin{array}{l}
    \textnormal{Mean}(F^\ast_{\bullet}, F^e_{\bullet}) = \max{(\textnormal{mean}(F^\ast_{\bullet,\,i,\,j},\,F^e_{\bullet,\,i,\,j}))} \\
    \textnormal{Mean}(F^e_{\bullet}, F^r_{\bullet}) =\max{(\textnormal{mean}(F^e_{\bullet,\,i,\,j},\,F^r_{\bullet,\,i,\,j}))} \\
    \textnormal{RMS}(F^\ast_{\bullet}, F^e_{\bullet})=\max{(\textnormal{rmse}(F^\ast_{\bullet,\,i,\,j},\,F^e_{\bullet,\,i,\,j}))} \\
    \textnormal{RMS}(F^e_{\bullet}, F^r_{\bullet})=\max{(\textnormal{rmse}(F^e_{\bullet,\,i,\,j},\,F^r_{\bullet,\,i,\,j}))} \\
    \textnormal{Max}(F^{\ast}_{\bullet}, F^r_{\bullet})=\max{(\max(F^\ast_{\bullet,\,i,\,j},\,F^e_{\bullet,\,i,\,j}))}\\
    \textnormal{Max}(F^e_{\bullet}, F^r_{\bullet})=\max{(\max(F^e_{\bullet,\,i,\,j},\,F^r_{\bullet,\,i,\,j}))}
    \end{array}
    \right.
\, ,
\end{align*}
where $i\in(1,\,2,\,3)$ denotes different $\theta = (10^\circ, \, 30^\circ, \, 50^\circ)$, $j\in(1,\,2,\,3,\,4,\,5)$ is the index of the five repeated experiments, and $\bullet$ denotes subscript \{$g$, $p$\} for grasping and pulling forces, respectively.

\subsection{Tissue Resection Operation Flow}

The operation flows of resections conducted in Fig.~\ref{Figure:exvivo}, Fig.~\ref{Figure:exvivo_without}, and Fig.~\ref{Figure:exvivo_beef} are summarized in Algorithm~\ref{Algorithm:operation_flow}.
\begin{algorithm}[h!]
	\begin{algorithmic}[1]
		\Require
		 $F_g^{\ast}$, $F_p^{\ast, \: 0}$, $\rho$, $\Delta d_p^t$, $d_p^t$, and $F^{\ast, \, c}_p$
		\While{$F_p \geq F_p^{\ast, \, t}$}  {\footnotesize~~~~~~~~~~~~~~~~~~~~~//\textit{Touch detection}}
		\State move forceps forward; 
		\EndWhile
		\While{$F_g \leq F_g^{\ast}$} {\footnotesize~~~~~~~~~~~~~~~~~~~~~~~//\textit{Tissue grasping}}
			\State grasp tissue with (\ref{equation:Fg_lower}) and (\ref{Equation:Mode-I});
		\EndWhile
		\While{$F_p \leq F_p^{\ast,\: 0}$} {\footnotesize~~~~~~~~~~~~~~~~~~~~//\textit{Tissue pulling}}
			\State pull tissue with (\ref{Equation:Mode-II}) and (\ref{Equation:Mode-I}); {\footnotesize//\textit{Exps in Fig.~\ref{Figure:exvivo}, \ref{Figure:exvivo_beef}}}
           \State // pull tissue with (\ref{Equation:Mode-II}) and (\ref{equation:Fp_upper}); {\footnotesize//\textit{Exp in Fig.~\ref{Figure:exvivo_without}}}
		\EndWhile
	
		\While{$F_p \ge F^{\ast, \, c}_p$} {\footnotesize~~~~~~~~// \textit{Tissue cutting and pulling}}
			\State cut tissue with scissors;
			\State $i = i+1$;
			\State $F_p^{\ast,\: i} = \rho F_p^{\ast,\: i-1}$;
			\While{$F_p \leq F_p^{\ast,\: i}$ and $d_p \leq d_p^t$}		
				\While{$\Delta d_p \leq \Delta d_p^t$}
        			\State pull tissue with (\ref{Equation:Mode-II}) and (\ref{Equation:Mode-I});{\footnotesize~//\textit{Exps in Fig.~\ref{Figure:exvivo}, \ref{Figure:exvivo_beef}}}	
                    \State // pull tissue with (\ref{Equation:Mode-II}) and (\ref{equation:Fp_upper});{\footnotesize~//\textit{Exp in Fig.~\ref{Figure:exvivo_without}}}
				\EndWhile		
			\EndWhile			
		\EndWhile
            \State final operator check and cut tissue if required;
		\State move out the cutoff tissue;
	\end{algorithmic}	
	\caption{Tissue Resection Operation Flow}
	\label{Algorithm:operation_flow}
\end{algorithm}


\bibliographystyle{IEEEtran}
\bibliography{AutonmousTissueCutting}

\begin{thebibliography}{10}
\providecommand{\url}[1]{#1}
\csname url@samestyle\endcsname
\providecommand{\newblock}{\relax}
\providecommand{\bibinfo}[2]{#2}
\providecommand{\BIBentrySTDinterwordspacing}{\spaceskip=0pt\relax}
\providecommand{\BIBentryALTinterwordstretchfactor}{4}
\providecommand{\BIBentryALTinterwordspacing}{\spaceskip=\fontdimen2\font plus
\BIBentryALTinterwordstretchfactor\fontdimen3\font minus
  \fontdimen4\font\relax}
\providecommand{\BIBforeignlanguage}[2]{{%
\expandafter\ifx\csname l@#1\endcsname\relax
\typeout{** WARNING: IEEEtran.bst: No hyphenation pattern has been}%
\typeout{** loaded for the language `#1'. Using the pattern for}%
\typeout{** the default language instead.}%
\else
\language=\csname l@#1\endcsname
\fi
#2}}
\providecommand{\BIBdecl}{\relax}
\BIBdecl

\bibitem{yang2018grand}
G.-Z. Yang, J.~Bellingham, P.~E. Dupont, P.~Fischer, L.~Floridi, R.~Full,
  N.~Jacobstein, V.~Kumar, M.~McNutt, R.~Merrifield \emph{et~al.}, ``The grand
  challenges of science robotics,'' \emph{Science robotics}, vol.~3, no.~14, p.
  eaar7650, 2018.

\bibitem{dupont2021decade}
P.~E. Dupont, B.~J. Nelson, M.~Goldfarb, B.~Hannaford, A.~Menciassi, M.~K.
  O’Malley, N.~Simaan, P.~Valdastri, and G.-Z. Yang, ``A decade retrospective
  of medical robotics research from 2010 to 2020,'' \emph{Science Robotics},
  vol.~6, no.~60, p. eabi8017, 2021.

\bibitem{dupont2022continuum}
P.~E. Dupont, N.~Simaan, H.~Choset, and C.~Rucker, ``Continuum robots for
  medical interventions,'' \emph{Proceedings of the IEEE}, vol. 110, no.~7, pp.
  847--870, 2022.

\bibitem{razjigaev2022end}
A.~Razjigaev, A.~K. Pandey, D.~Howard, J.~Roberts, and L.~Wu, ``End-to-end
  design of bespoke, dexterous snake-like surgical robots: A case study with
  the raven ii,'' \emph{IEEE Transactions on Robotics}, vol.~38, no.~5, pp.
  2827--2840, 2022.

\bibitem{nwafor2023design}
C.~J. Nwafor, C.~Girerd, G.~J. Laurent, T.~K. Morimoto, and K.~Rabenorosoa,
  ``Design and fabrication of concentric tube robots: A survey,'' \emph{IEEE
  Transactions on Robotics}, 2023.

\bibitem{wu2017development}
L.~Wu, S.~Song, K.~Wu, C.~M. Lim, and H.~Ren, ``Development of a compact
  continuum tubular robotic system for nasopharyngeal biopsy,'' \emph{Medical
  \& biological engineering \& computing}, vol.~55, pp. 403--417, 2017.

\bibitem{wang2023novel}
J.~Wang, C.~Hu, G.~Ning, L.~Ma, X.~Zhang, and H.~Liao, ``A novel miniature
  spring-based continuum manipulator for minimally invasive surgery: Design and
  evaluation,'' \emph{IEEE/ASME Transactions on Mechatronics}, 2023.

\bibitem{hu2023design}
Z.~Hu, J.~Li, and S.~Wang, ``Design and kinematics of a robotic instrument for
  natural orifice transluminal endoscopic surgery,'' \emph{IEEE/ASME
  Transactions on Mechatronics}, 2023.

\bibitem{price2023using}
K.~Price, J.~Peine, M.~Mencattelli, Y.~Chitalia, D.~Pu, T.~Looi, S.~Stone,
  J.~Drake, and P.~E. Dupont, ``Using robotics to move a neurosurgeon’s hands
  to the tip of their endoscope,'' \emph{Science Robotics}, vol.~8, no.~82, p.
  eadg6042, 2023.

\bibitem{gao2023transendoscopic}
H.~Gao, X.~Yang, X.~Xiao, X.~Zhu, T.~Zhang, C.~Hou, H.~Liu, M.~Q.-H. Meng,
  L.~Sun, X.~Zuo \emph{et~al.}, ``Transendoscopic flexible parallel continuum
  robotic mechanism for bimanual endoscopic submucosal dissection,'' \emph{The
  International Journal of Robotics Research}, p. 02783649231209338, 2023.

\bibitem{wu2022camera}
L.~Wu, F.~Yu, T.~N. Do, and J.~Wang, ``Camera frame misalignment in a
  teleoperated eye-in-hand robot: Effects and a simple correction method,''
  \emph{IEEE Transactions on Human-Machine Systems}, vol.~53, no.~1, pp. 2--12,
  2022.

\bibitem{kong2022design}
Y.~Kong, S.~Song, N.~Zhang, J.~Wang, and B.~Li, ``Design and kinematic modeling
  of in-situ torsionally-steerable flexible surgical robots,'' \emph{IEEE
  Robotics and Automation Letters}, vol.~7, no.~2, pp. 1864--1871, 2022.

\bibitem{yu2016development}
H.~Yu, L.~Wu, K.~Wu, and H.~Ren, ``Development of a multi-channel concentric
  tube robotic system with active vision for transnasal nasopharyngeal
  carcinoma procedures,'' \emph{IEEE Robotics and Automation Letters}, vol.~1,
  no.~2, pp. 1172--1178, 2016.

\bibitem{li2023design}
Z.~Li, X.~Li, J.~Lin, D.~Yang, J.~Xu, S.~Zhang, and J.~Guo, ``Design and
  application of multi-dimensional force/torque sensors in surgical robots: A
  review,'' \emph{IEEE Sensors Journal}, 2023.

\bibitem{van2009value}
O.~A. Van~der Meijden and M.~P. Schijven, ``The value of haptic feedback in
  conventional and robot-assisted minimal invasive surgery and virtual reality
  training: a current review,'' \emph{Surgical endoscopy}, vol.~23, pp.
  1180--1190, 2009.

\bibitem{khadem2016modular}
S.~M. Khadem, S.~Behzadipour, A.~Mirbagheri, and F.~Farahmand, ``A modular
  force-controlled robotic instrument for minimally invasive surgery--efficacy
  for being used in autonomous grasping against a variable pull force,''
  \emph{The International Journal of Medical Robotics and Computer Assisted
  Surgery}, vol.~12, no.~4, pp. 620--633, 2016.

\bibitem{abiri2019multi}
A.~Abiri, J.~Pensa, A.~Tao, J.~Ma, Y.-Y. Juo, S.~J. Askari, J.~Bisley,
  J.~Rosen, E.~P. Dutson, and W.~S. Grundfest, ``Multi-modal haptic feedback
  for grip force reduction in robotic surgery,'' \emph{Scientific reports},
  vol.~9, no.~1, p. 5016, 2019.

\bibitem{skandalakis2021surgical}
J.~E. Skandalakis, P.~N. Skandalakis, and L.~J. Skandalakis, ``Thyroglossal
  duct cystectomy,'' in \emph{Surgical anatomy and technique}.\hskip 1em plus
  0.5em minus 0.4em\relax Springer, 2021, pp. 69--72.

\bibitem{liu2022recent}
Z.~Liu, Z.~Liao, D.~Wang, C.~Wang, C.~Song, H.~Li, and Y.~Liu, ``Recent
  advances in soft biological tissue manipulating technologies,'' \emph{Chinese
  Journal of Mechanical Engineering}, vol.~35, no.~1, pp. 1--34, 2022.

\bibitem{kim2018sensorized}
U.~Kim, Y.~B. Kim, J.~So, D.-Y. Seok, and H.~R. Choi, ``Sensorized surgical
  forceps for robotic-assisted minimally invasive surgery,'' \emph{IEEE
  Transactions on Industrial Electronics}, vol.~65, no.~12, pp. 9604--9613,
  2018.

\bibitem{seok2019compensation}
D.-Y. Seok, Y.~B. Kim, U.~Kim, S.~Y. Lee, and H.~R. Choi, ``Compensation of
  environmental influences on sensorized-forceps for practical surgical
  tasks,'' \emph{IEEE Robotics and Automation Letters}, vol.~4, no.~2, pp.
  2031--2037, 2019.

\bibitem{liu2023hapticsenabled}
T.~Liu, T.~Zhang, J.~Katupitiya, J.~Wang, and L.~Wu, ``Haptics-enabled forceps
  with multimodal force sensing: Toward task-autonomous surgery,''
  \emph{IEEE/ASME Transactions on Mechatronics}, vol.~29, no.~3, pp.
  2208--2219, 2024.

\bibitem{nazari2021image}
A.~A. Nazari, F.~Janabi-Sharifi, and K.~Zareinia, ``Image-based force
  estimation in medical applications: A review,'' \emph{IEEE Sensors Journal},
  vol.~21, no.~7, pp. 8805--8830, 2021.

\bibitem{haouchine2018vision}
N.~Haouchine, W.~Kuang, S.~Cotin, and M.~Yip, ``Vision-based force feedback
  estimation for robot-assisted surgery using instrument-constrained
  biomechanical three-dimensional maps,'' \emph{IEEE Robotics and Automation
  Letters}, vol.~3, no.~3, pp. 2160--2165, 2018.

\bibitem{lee2018interaction}
D.-H. Lee, W.~Hwang, and S.-C. Lim, ``Interaction force estimation using camera
  and electrical current without force/torque sensor,'' \emph{IEEE Sensors
  Journal}, vol.~18, no.~21, pp. 8863--8872, 2018.

\bibitem{li2020super}
Y.~Li, F.~Richter, J.~Lu, E.~K. Funk, R.~K. Orosco, J.~Zhu, and M.~C. Yip,
  ``Super: A surgical perception framework for endoscopic tissue manipulation
  with surgical robotics,'' \emph{IEEE Robotics and Automation Letters},
  vol.~5, no.~2, pp. 2294--2301, 2020.

\bibitem{nwoye2022artificial}
E.~Nwoye, W.~L. Woo, B.~Gao, and T.~Anyanwu, ``Artificial intelligence for
  emerging technology in surgery: Systematic review and validation,''
  \emph{IEEE Reviews in Biomedical Engineering}, 2022.

\bibitem{lanir2017multi}
Y.~Lanir, ``Multi-scale structural modeling of soft tissues mechanics and
  mechanobiology,'' \emph{Journal of Elasticity}, vol. 129, pp. 7--48, 2017.

\bibitem{bandari2019tactile}
N.~Bandari, J.~Dargahi, and M.~Packirisamy, ``Tactile sensors for minimally
  invasive surgery: A review of the state-of-the-art, applications, and
  perspectives,'' \emph{Ieee Access}, vol.~8, pp. 7682--7708, 2019.

\bibitem{king2009tactile}
C.-H. King, M.~O. Culjat, M.~L. Franco, C.~E. Lewis, E.~P. Dutson, W.~S.
  Grundfest, and J.~W. Bisley, ``Tactile feedback induces reduced grasping
  force in robot-assisted surgery,'' \emph{IEEE transactions on haptics},
  vol.~2, no.~2, pp. 103--110, 2009.

\bibitem{lai2021three}
W.~Lai, L.~Cao, J.~Liu, S.~C. Tjin, and S.~J. Phee, ``A three-axial force
  sensor based on fiber bragg gratings for surgical robots,'' \emph{IEEE/ASME
  Transactions on Mechatronics}, vol.~27, no.~2, pp. 777--789, 2021.

\bibitem{zarrin2018development}
P.~S. Zarrin, A.~Escoto, R.~Xu, R.~V. Patel, M.~D. Naish, and A.~L. Trejos,
  ``Development of a 2-dof sensorized surgical grasper for grasping and axial
  force measurements,'' \emph{IEEE Sensors Journal}, vol.~18, no.~7, pp.
  2816--2826, 2018.

\bibitem{kim2015force}
U.~Kim, D.-H. Lee, W.~J. Yoon, B.~Hannaford, and H.~R. Choi, ``Force sensor
  integrated surgical forceps for minimally invasive robotic surgery,''
  \emph{IEEE Transactions on Robotics}, vol.~31, no.~5, pp. 1214--1224, 2015.

\bibitem{xiang2023learning}
P.~Xiang, J.~Zhang, D.~Sun, K.~Qiu, Q.~Fang, X.~Mi, Y.~Wang, R.~Xiong, and
  H.~Lu, ``Learning-based high-precision force estimation and compliant control
  for small-scale continuum robot,'' \emph{IEEE Transactions on Automation
  Science and Engineering}, 2023.

\bibitem{wijayarathne2023real}
L.~Wijayarathne, Z.~Zhou, Y.~Zhao, and F.~L. Hammond, ``Real-time
  deformable-contact-aware model predictive control for force-modulated
  manipulation,'' \emph{IEEE Transactions on Robotics}, 2023.

\bibitem{sheng2021hybrid}
Q.~Sheng, Z.~Geng, L.~Hua, and X.~Sheng, ``Hybrid vision-force robot force
  control for tasks on soft tissues,'' in \emph{2021 27th International
  Conference on Mechatronics and Machine Vision in Practice (M2VIP)}.\hskip 1em
  plus 0.5em minus 0.4em\relax IEEE, 2021, pp. 705--710.

\bibitem{stoll2006force}
J.~Stoll and P.~Dupont, ``Force control for grasping soft tissue,'' in
  \emph{Proc. of IEEE Int. Conf. of Robotics and Automation}, 2006, pp.
  4309--4311.

\bibitem{yu2007comparison}
X.~Yu, H.~J. Chizeck, and B.~Hannaford, ``Comparison of transient performance
  in the control of soft tissue grasping,'' in \emph{2007 IEEE/RSJ
  International Conference on Intelligent Robots and Systems}.\hskip 1em plus
  0.5em minus 0.4em\relax IEEE, 2007, pp. 1809--1814.

\bibitem{waters2022incipient}
I.~Waters, L.~Wang, D.~Jones, A.~Alazmani, and P.~Culmer, ``Incipient slip
  sensing for improved grasping in robot assisted surgery,'' \emph{IEEE Sensors
  Journal}, vol.~22, no.~16, pp. 16\,545--16\,554, 2022.

\bibitem{chua2022characterization}
Z.~Chua and A.~M. Okamura, ``Characterization of real-time haptic feedback from
  multimodal neural network-based force estimates during teleoperation,'' in
  \emph{2022 IEEE/RSJ International Conference on Intelligent Robots and
  Systems (IROS)}.\hskip 1em plus 0.5em minus 0.4em\relax IEEE, 2022, pp.
  1471--1478.

\bibitem{mathur2019evaluation}
B.~Mathur, A.~Topiwala, H.~Saeidi, T.~Fleiter, and A.~Krieger, ``Evaluation of
  control strategies for a tele-manipulated robotic system for remote trauma
  assessment,'' in \emph{2019 Proceedings of the Conference on Control and its
  Applications}.\hskip 1em plus 0.5em minus 0.4em\relax SIAM, 2019, pp. 7--14.

\bibitem{ang2005pid}
K.~H. Ang, G.~Chong, and Y.~Li, ``Pid control system analysis, design, and
  technology,'' \emph{IEEE transactions on control systems technology},
  vol.~13, no.~4, pp. 559--576, 2005.

\bibitem{woo2021fluorescent}
Y.~Woo, S.~Chaurasiya, M.~O’Leary, E.~Han, and Y.~Fong, ``Fluorescent imaging
  for cancer therapy and cancer gene therapy,'' \emph{Molecular
  Therapy-Oncolytics}, vol.~23, pp. 231--238, 2021.

\bibitem{HuiNip2001BeefTexture}
Y.~H. Hui and W.-K. Nip, \emph{Beef Texture and Juiciness}.\hskip 1em plus
  0.5em minus 0.4em\relax New York: CRC Press, 2001, ch.~9, pp. 177--206.

\end{thebibliography}
\begin{IEEEbiography}[{\includegraphics[width=1in,height=1.25in,clip,keepaspectratio]{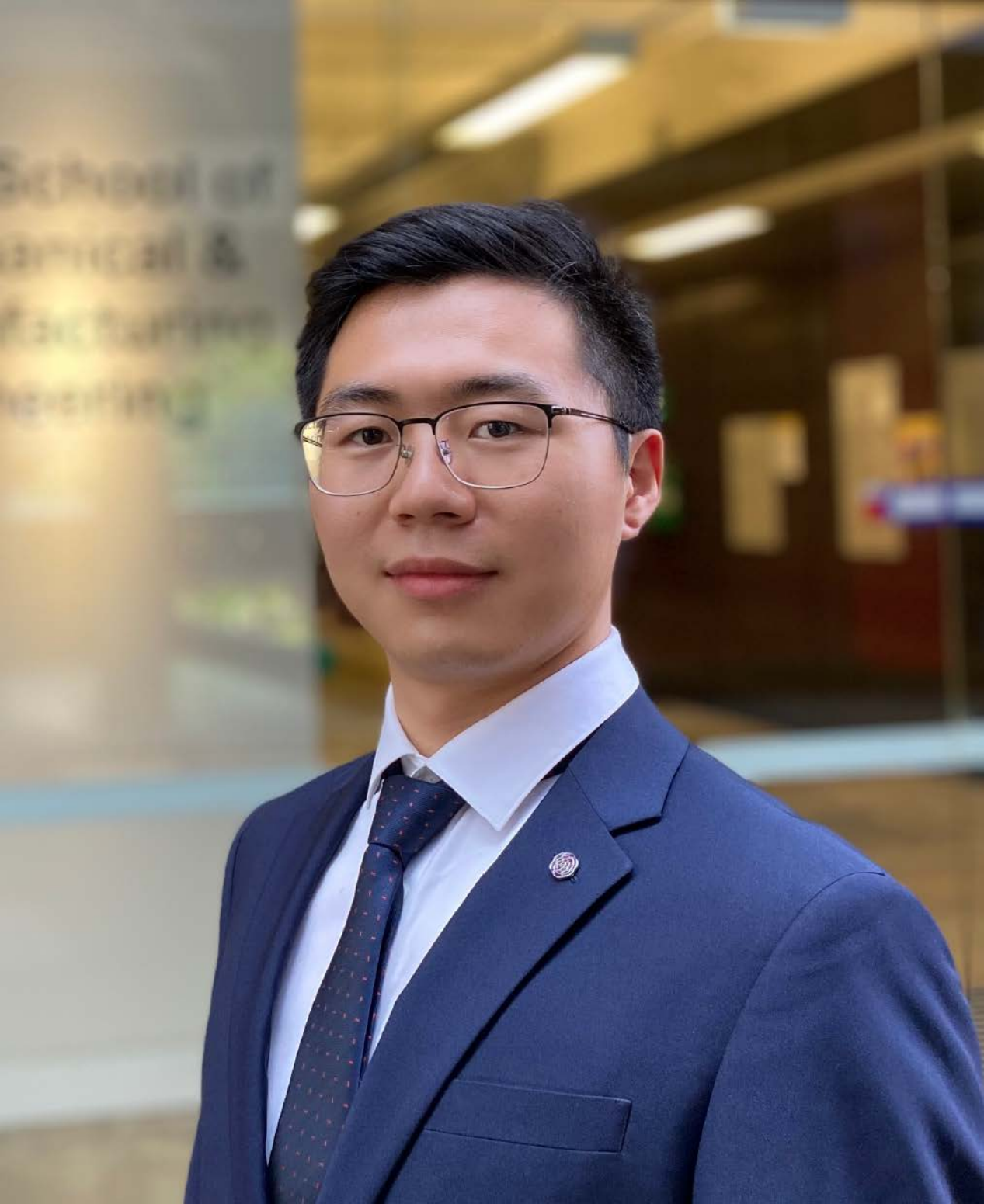}}] {Tangyou Liu} received the B.E. degree in mechanical engineering from the Southwest University of Science and Technology (SWUST), Mianyang, China, in 2017, and the M.E. degree in mechanical engineering,
as an outstanding graduate, from the Harbin Institute of Technology, Shenzhen (HITsz), Shenzhen, China, in 2021, supervised by Prof. Max~Q.-H.~Meng. He is currently working toward the Ph.D. degree in mechatronic engineering with the University of New South Wales (UNSW), Sydney, NSW, Australia, under the supervision of Dr. Liao Wu. 
He was with KUKA, China, as a System Development Engineer.
He was awarded the champion of the KUKA China R\&D Innovation Challenge and the IEEE ICRA2023 Best Poster.
His current research interests include medical and surgical robots.
\end{IEEEbiography}

\begin{IEEEbiography}[{\includegraphics[width=1in,height=1.25in,clip,keepaspectratio]{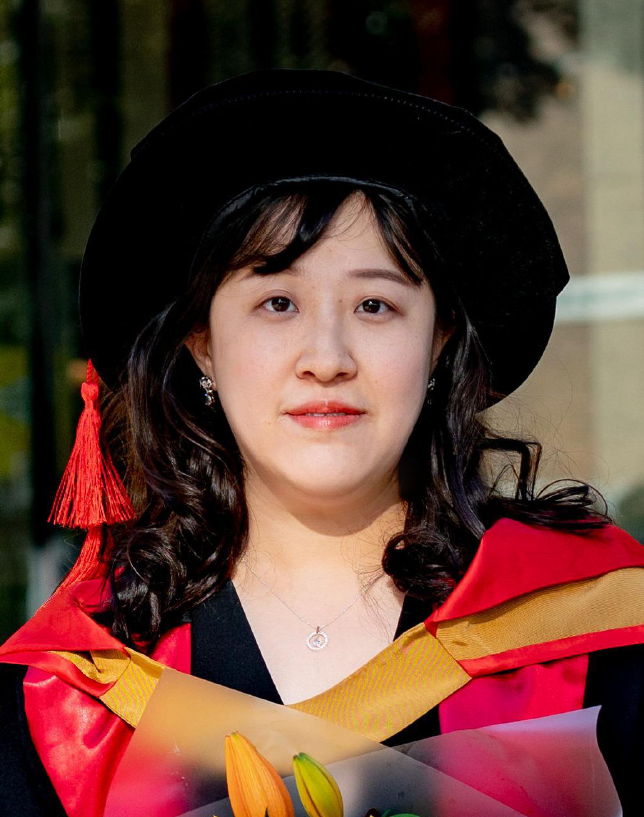}} ] {Xiaoyi Wang} received the B.E. degree in Automation and M.E. degree in Control Engineering from Harbin Institute of Technology (HIT), Harbin, China in 2014 and 2017, respectively. Then, she received her Ph.D. degree from University of New South Wales (UNSW), Sydney, Australia, in 2022, where she is working as a postdoctoral research fellow.
Her current research interests include the control of aerospace robots and medical robots, nonlinear control, and stochastic optimal control.
\end{IEEEbiography}

\begin{IEEEbiography}[{\includegraphics[width=1in,height=1.25in,clip,keepaspectratio]{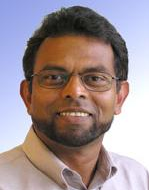}}]{Jay Katupitiya} received the B.S. degree in
production engineering from the University of Peradeniya, Peradeniya, Sri Lankain, in 1978, and the Ph.D. degree in mechanical engineering from
the Katholieke Universiteit Leuven, Leuven, Belgium, in 1985.
He is currently an Associate Professor with the University of New South Wales, Sydney, NSW, Australia. His main research interest includes the development of sophisticated control methodologies for the precision guidance of field vehicles on rough terrain at high speeds.
\end{IEEEbiography}

\begin{IEEEbiography}[{\includegraphics[width=1in,height=1.25in,clip,keepaspectratio]{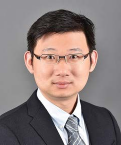}}]{Jiaole Wang} received the B.E. degree in mechanical engineering from Beijing Information Science and Technology University,
Beijing, China, in 2007, the M.E. degree in human \& artificial intelligent systems from the Department of Human and Artificial Intelligent Systems, University of Fukui, Fukui, Japan, in 2010, and the Ph.D. degree in electronic engineering from the Department of Electronic Engineering, The Chinese University of Hong Kong (CUHK), Hong Kong, in 2016.
He was a Research Fellow with Pediatric Cardiac Bioengineering Laboratory, Department of Cardiovascular Surgery, Boston Children’s Hospital and Harvard Medical School, Boston, MA, USA. He is currently an Associate Professor with the School of Biomedical Engineering and Digital Health, Harbin Institute of Technology (Shenzhen), Shenzhen, China. His main research interests include medical and surgical robotics, image-guided surgery, human-robot interaction, and magnetic tracking and actuation for biomedical applications.
\end{IEEEbiography}
\newpage

\begin{IEEEbiography}[{\includegraphics[width=1in,height=1.25in,clip,keepaspectratio]{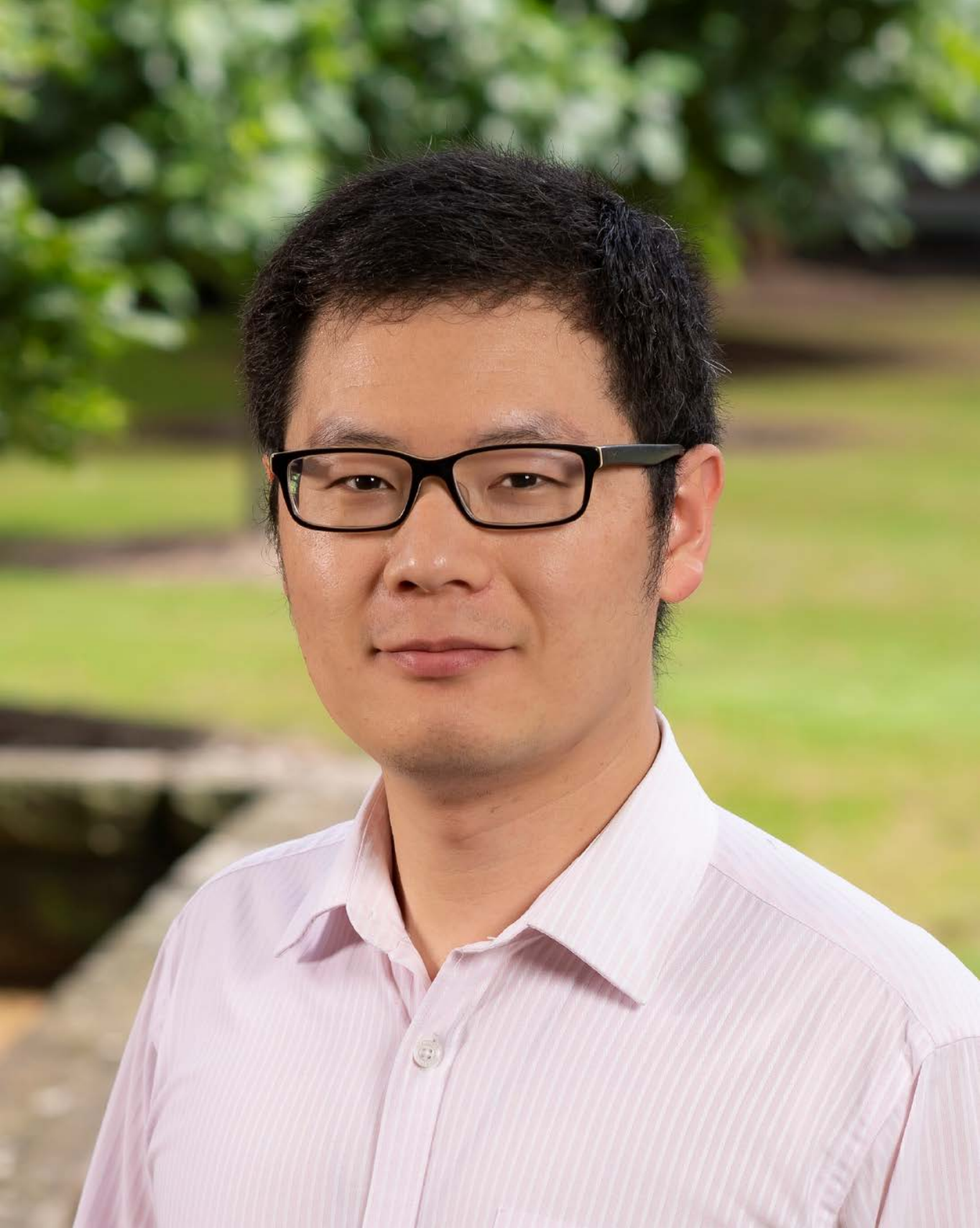}}]{Liao Wu} received the B.S. and Ph.D. degrees in mechanical engineering from Tsinghua University, Beijing, China, in 2008 and 2013, respectively.
From 2014 to 2015, he was a Research Fellow with National University of Singapore, Singapore. From 2016 to 2018, he was a Vice-Chancellor’s Research Fellow with Queensland University of Technology, Brisbane, Australia. 
From 2016 to 2020, he was with the Australian Centre for Robotic Vision, an ARC Centre of Excellence. 
He is currently a Senior Lecturer with the School of Mechanical and Manufacturing Engineering, University of New South Wales, Sydney, Australia. His research interests cover medical and industrial robotics, including flexible and intelligent surgical robot design, human-robot interaction, assistive robotics, and robot modeling and calibration.
\end{IEEEbiography}
\end{document}